%% file: main.tex
\documentclass[times,twocolumn,final,authoryear]{elsarticle}

\usepackage[T1]{fontenc}
\usepackage[utf8]{inputenc}

\usepackage{ycviu}
\usepackage{framed,multirow}

\usepackage{latexsym}

\usepackage{xcolor}
\usepackage[colorlinks=true]{hyperref}

\definecolor{DarkGreen}{rgb}{0.2,0.5,0.2}

\usepackage{microtype}

\usepackage{graphicx}
\usepackage{amsmath}
\usepackage{amssymb}
\usepackage{physics}
\usepackage{booktabs}
\usepackage{multirow}

\usepackage{setspace}

\usepackage[ruled,vlined]{algorithm2e}

\usepackage{pifont}

\usepackage{bm}
\usepackage{caption}

\usepackage{wrapfig}
\usepackage{multirow}

\usepackage{pdfpages}

\usepackage{balance}

\usepackage{stfloats}

\usepackage{fancyhdr} 
\newcommand{\bv}[1]{\mathbf{#1}}
\newcommand{\mI}{\boldsymbol{\mathcal{I}}}
\newcommand{\mD}{\boldsymbol{\mathcal{D}}}
\newcommand{\mK}{\boldsymbol{\mathcal{K}}}
\newcommand{\mF}{\boldsymbol{\mathcal{F}}}

\newcommand{\mV}{\boldsymbol{\mathcal{V}}}
\newcommand{\mE}{\boldsymbol{\mathcal{E}}}
\newcommand{\mM}{\boldsymbol{\mathcal{M}}}

\usepackage[colorinlistoftodos]{todonotes}

\newcommand{\new}[1]{{{#1}}}%

\newcommand{\wrt}{{w.r.t.} }
\newcommand{\ie}{\textit{i}.\textit{e}.}

\journal{Computer Vision and Image Understanding}

\begin{document}

\ifpreprint
  \setcounter{page}{1}
\else
  \setcounter{page}{1}
\fi

\begin{frontmatter}

\title{Learning Geodesic-Aware Local Features from RGB-D Images}

\author[1]{Guilherme Potje \corref{cor1}} 
\cortext[cor1]{Corresponding author: 
}
\ead{guipotje@dcc.ufmg.br}
\author[2]{Renato Martins}
\author[1]{Felipe Cadar}
\author[1]{Erickson R. Nascimento}

\address[1]{Department of Computer Science -- Universidade Federal de Minas Gerais, Belo Horizonte, Brazil.}
\address[2]{VIBOT EMR CNRS 6000, ImViA, Université Bourgogne Franche-Comté, Le Creusot, France.}

\begin{abstract}
	Most of the existing handcrafted and learning-based local descriptors are still at best approximately invariant to affine image transformations, often disregarding deformable surfaces. In this paper, we take one step further by proposing a new approach to compute descriptors from RGB-D images (where RGB refers to the pixel color brightness and D stands for depth information) that are invariant to isometric non-rigid deformations, as well as to scale changes and rotation. Our proposed description strategies are grounded on the key idea of learning feature representations on undistorted local image patches using surface geodesics. We design two complementary local descriptors strategies to compute geodesic-aware features efficiently: one efficient binary descriptor based on handcrafted binary tests (named GeoBit), and one learning-based descriptor (GeoPatch) with convolutional neural networks (CNNs) to compute features. In different experiments using real and publicly available RGB-D data benchmarks, they consistently outperforms state-of-the-art handcrafted and learning-based image and RGB-D descriptors in matching scores, as well as in object retrieval and non-rigid surface tracking experiments, with comparable processing times. 
	We also provide to the community a new dataset with accurate matching annotations of RGB-D images of different objects (shirts, cloths, paintings, bags), subjected to strong non-rigid deformations, for evaluation benchmark of deformable surface correspondence algorithms.
	
\end{abstract}


\end{frontmatter}



\input{introduction}

\input{related_work}

\input{methodology}

\input{experiments}

\input{tracking}

\input{conclusion}

\bibliographystyle{model2-names}

\input{main.bbl}
\input{appendix}

\end{document}

%% file: introduction.tex
\section{Introduction}

\begin{figure}[t!]
	\centering
	\includegraphics[width=0.95\columnwidth]{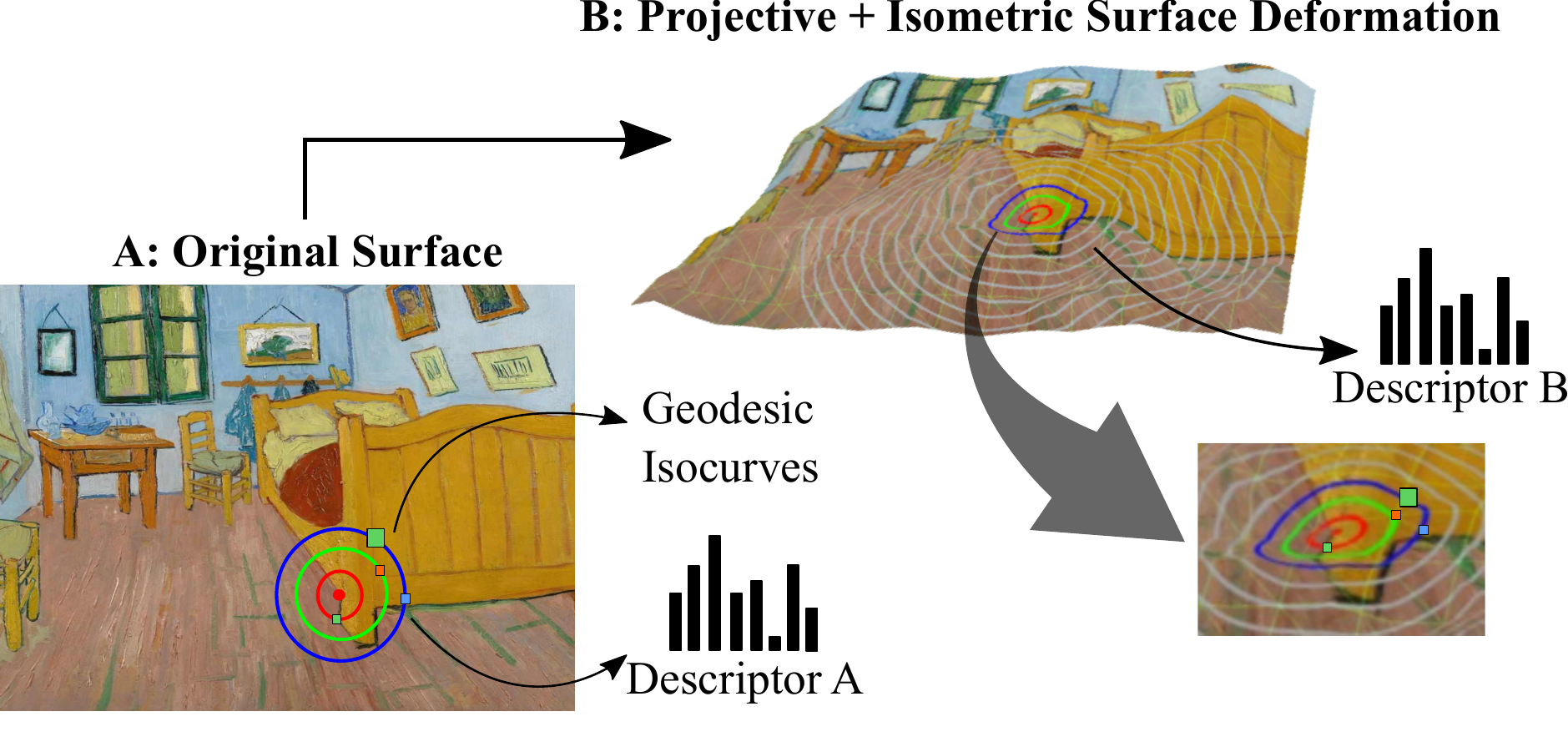}
	\caption{\textbf{Extracting geodesic features}. Due to the invariance property of the geodesic distance, the pixel intensities lying on the same isocurves do not change after the deformations and can be sampled to compute an invariant descriptor. Descriptors A and B are still similar (pixels indicated by squares in red, blue, and green lie on the same isocurve after the deformation).}\label{fig:teaserS-geodesic-aware}	
\end{figure}
Finding accurate correspondences between images plays a key role in the development of solutions for a wide range of computer vision problems. Descriptors are meant to provide distinguished representations to characterize visual information from the world ubiquitously and relate images taken from different locations, illuminations, and perspectives. Over the past few decades, the computer vision community has witnessed the rise of  handcrafted~\citep{lowe2004ijcv,daisy,rublee2011iccv, brief2012tpami} and learning-based local descriptors~\citep{cit:LIFT,tfeat-balntas2016,cit:l2net,cit:hardnet,cit:beyondcartesian} for image correspondence. Despite remarkable advances in describing local patches in images, their full integration in solutions for real-world applications still faces several challenges. 
For instance, while visually recognizing or tracking objects, we need to deal with severe conditions as partial view occlusions, rotations, and illumination changes, but also with the challenging condition of non-rigid surface deformations.

Different local feature descriptors for intensity images~\citep{lowe2004ijcv,daisy, rublee2011iccv,brief2012tpami} have been proposed since images usually contain a rich source of information from the scenes and objects' characteristics. 
Even though image-based methods tend to exploit much of the rich information engrafted in images wisely, these techniques are restricted to 2D data. Thus, the performance of texture-based descriptors tends to further degrade when non-rigid deformations arise in the scene. Moreover, scale and rotation invariance in these descriptors are usually achieved by an a priori estimation of the scale and orientation parameters, which is often noisy and ambiguous. Geometric information such as depth images, for its turn, is becoming increasingly popular for computing distinctive feature descriptors. Geometric features are less sensitive to the lack of texture and allow complementary description of features on surfaces.
Multimodal approaches~\citep{zaharescu2009cvpr,tombari2011icip,nascimento2012iros,nascimento2012icpr} have shown that the integration of both the rich texture information from intensity images with geometry cues from depth maps can improve the description of sparse keypoints.

Despite the successful use of the methods as mentioned earlier on several problems, they are approximately invariant to affine transformations and local illumination changes. When strong affine or non-rigid deformations arise in the image, their performance might drastically decrease. Since in real-world applications, objects may assume different forms when being bent or folded, other types of transformations are worth considering. Isometry, for instance, is a non-rigid transformation that usually appears in the world around us. An isometric deformation is a length-preserving mapping, where the geodesic distance does not change in the manifold after deforming the surface.  \figurename~\ref{fig:teaserS-geodesic-aware} illustrates the isometric deformation and the geodesic invariance concepts applied in the context of local visual feature extraction. 
Several real-world objects are affected by non-rigid isometric deformations, i.e., surfaces or objects that can bend and resist stretching (and contractions). Typical examples are cloth, material surfaces as paper and articulated objects. In order to unambiguously and reliably represent and describe these objects, we need to compute invariant features of the object's under these non-rigid deformations. Therefore, non-rigid tracking and reconstruction~\citep{bozic2020deepdeform} are tasks that can directly profit from methods that robustly handle correspondences of surfaces under deformation.

Some existing keypoint description approaches approximate geometric intrinsic cues solely from image intensity, such as done in DaLI descriptor~\citep{dali}, however they suffer from high computational costs and loss of distinctiveness in exchange for its robustness to deformations. Similarly, geometric-aware methods that disregard the non-rigid nature of objects like BRAND~\citep{nascimento2012iros}, are also greatly penalized when deformation arises in the scene, as can be observed in our experimental section. In this paper, we take a step towards local feature extraction invariance to non-rigid deformations in RGB-D images. In the experiments, we employed our descriptors to track and retrieve objects demonstrating its effectiveness in handling strong isometric deformations beyond different scales and rotations. The main contributions of this paper can be summarized as follows: i) We extend the binary descriptor GeoBit, previously introduced by \cite{nascimento2019iccv}, by incorporating a more efficient strategy for estimating geodesic distances around a keypoint and demonstrate the improvement in the feature representation; ii) We propose a new learning-based descriptor, called GeoPatch, that outperforms other methods in accuracy and is competitive in terms of processing time.

We show how to leverage the equivariance rotation property of the polar sampling for training a network to attain rotation invariance without explicitly learning any additional task;

iii) We also provide to the community a new RGB-D benchmark comprising $11$ real-world objects and a large simulated dataset of RGB-D images. All real-world objects, captured with Kinect{\texttrademark}, versions 1 and 2, were subjected to a variety of non-rigid deformations, with ground-truth matches obtained by manual annotation and an accurate motion capture system. Dense set of keypoints with correspondence is provided for all sequences by a thin-plate-spline deformation model. The implementation of our descriptor and simulation framework will be released, allowing the creation of new datasets and encouraging future research in developing invariant descriptors. 

%% file: related_work.tex
\section{Related Work}\label{sec:related}

Extracting descriptors from images usually provide rich information on the object features, while geometric images, produced by 3D sensors, are less sensitive to lighting conditions. Two representative approaches to images are SIFT~\citep{lowe2004ijcv} and SURF~\citep{surf2008cviu} descriptors. These descriptors first extract features using local gradients and then estimate a dominant orientation of the keypoint's neighborhood to provide invariance to rotation transformations.
Inspired by the idea of Local Binary Patterns (LBP)~\citep{ojala1996patternrecognition}, the use of binary strings to assembly the feature vector has become popular, and several binary descriptors such as BRIEF~\citep{brief2012tpami}, ORB~\citep{rublee2011iccv}, and BRISK~\citep{brisk11} have been proposed. As advantage, binary strings bring small computational cost and reduced storage requirements. Although being able to obtain invariance to affine transformations, these image-based descriptors are limited to rigid scenes.



One of the enduring grand challenges in shape analysis is to extract properties that preserve the intrinsic geometry of shapes. Geodesic distance is a well known intrinsic property as far as isometric transformations are concerned. The intrinsic shape context (ISC)~\citep{Kokkinos2012cvpr} descriptor is based on properties of the geodesic distance. \cite{Shamai2017GeodesicDD} proposed and evaluated a new basis for geodesic distance representation. The authors also showed how to approximate the geodesic distance efficiently. Despite advances achieved by these works, their technique is most suitable to 3D shapes only. In the same direction, BRISKS~\citep{brisks17}, a geodesic-aware BRISK descriptor targeted to spherical images is proposed. Similarly, SPHORB~\citep{sphorb} presents a binary descriptor based on ORB for spherical images. Both BRISKS and SPHORB, however, are designed to tackle solely 2-sphere manifolds, different from our descriptors that consider general and dynamic isometric image deformations.
 
A few studies have addressed resolving the problem of matching image keypoints on deformable surfaces. A representative approach that faces this problem includes the Deformation and Light Invariant (DaLI)~\citep{simo2015dali} descriptor. The authors proposed a new framework to use kernels based on diffusion geometry on 2D local patches. DaLI was designed to handle non-rigid image deformations and illumination changes. 
Despite remarkable advances in extracting features invariant to non-rigid image deformations, we show in our experiments that our approach outperforms DaLI in both quantitative metrics and computational effort in several experiments. 

The use of multiple cues, such as texture and geometric features, has shown to be an effective approach to improve the solution of several computer vision tasks such as semantic segmentation~\citep{huang2020pami,nakajima2019iccv}, mapping~\citep{martins2016adaptive,cavallari2020pami}, and the matching quality as well as to increase the discrimination power of feature vectors. \cite{martins2016adaptive} exploited the complementary properties of color and depth information encoded on RGB-D images to improve the convergence of direct (appearance-based) RGB-D registration. Another descriptor that uses both depth and texture is MeshHOG~\citep{zaharescu2009cvpr}. The authors used a texture extracted from 3D models to create scalar functions defined over a 2D manifold. CSHOT~\citep{tombari2011icip} exploits both texture and geometric features to create a local descriptor. 
\cite{nascimento2012iros, Nascimento:Neurocomputing:2013} presented BRAND, a lightweight local descriptor that encodes information as a binary string embedding geometric and texture cues and provides rotation and scale invariance. The fusion of depth and visual data was also exploited by \cite{liang2018cviu} to compute perspective invariant feature patches. Detectors of keypoint making use of both visual and geometrical information also have been proposed. \cite{vasconcelos2017prl} presented KVD, a keypoint detector that applies a decision tree to fuse depth and RGB data and enable their approach to work in the absence of visual data.

The previously mentioned works are carefully handcrafted designed methods for feature extraction, which was the norm until recently, except for some hybrid methods that involved some learning and optimization strategy~\citep{cit:sift_learned_winder, cit:brown_patches_learning}. The advent of deep learning has changed this scenario. Recent works based on convolutional networks showed the potential of learning descriptors from RGB images in an end-to-end manner \citep{tfeat-balntas2016, simo-iccv15, 3dmatch-cvpr17, wang-cvpr19}. Remarkable examples are LIFT~\citep{cit:LIFT}, SuperPoint~\citep{cit:superpoint}, and LF-Net~\citep{cit:lfnet}, where both the detection and description of keypoints are learned from data. Although these methods provide good results in images similar to the training set domain, when more general scenes arise, their performance tends to decrease. The authors of R2D2~\citep{revaud2019neurips} claim that some keypoints can be very salient but not so discriminative, and propose to jointly train a detector and descriptor based on the idea of predicting the distinctiveness of the detected keypoints, leading to a reliable keypoint detection that can be described confidently. The authors of Key.Net~\citep{cit:key.net} observed that using both handcrafted and learned CNN filters could outperform purely learned methods, since the handcrafted features can provide anchor filters, decreasing the complexity of the network, thus, increasing generalization. Following this direction, \cite{cit:beyondcartesian} demonstrated that training a CNN in a polar sampled patch can provide state-of-the-art results. Learned binary descriptors such as UDBD~\citep{cit:wu2020aggregation} demonstrate that it is possible to learn transformation-invariant binary codes from image patches for improved matching and retrieval. 


%% file: methodology.tex
\section{Methodology} \label{sec:method}

In this section, we describe the extraction of isometric-invariant visual features from RGB-D images. Our approach efficiently computes isometric invariant image patches from noisy RGB-D data by modeling the surface as a smooth 2D manifold and then estimating geodesic isocurves.  
Our strategy comprises two main steps: firstly, the geodesic distance computed from depth maps provides an isometric invariant mapping function $f(.)$. This mapping function returns the pixel location and retains invariance over deformations (as illustrated in the example shown in \figurename~\ref{fig:geodesic_mapping_function}); secondly, pixel intensities in images are used to extract distinctive features that will compose the descriptor feature vector $\mathbf{d}_f$.


We consider as input: a list of $l$ detected keypoints $\mK \in \mathbb{R}^{l\times 2} $; an estimate of the intrinsic camera matrix $\mathbf{K} \in \mathbb{R}^{3\times 3}$ and an RGB-D image encoding pixel color brightness and depth information $\mF = \{\mI,~\mD\}$, composed of a grayscale image $\mI \in [0,~1]^{H\times W}$ as pixel intensities and $\mD \in \mathbb{R}_+^{H\times W}$ as depth information. Thus, for each pixel $\tilde{\bv{p}} \in \mathbb{R}^2$ in Cartesian coordinates, $\mI(\tilde{\bv{p}})$ provides the pixel intensity and $\mD(\tilde{\bv{p}})$ the respective depth.

Please notice that the list of the detected keypoints $l$ can be computed with any existing keypoint detection algorithm, such as SIFT~\citep{lowe2004ijcv}, ORB~\citep{rublee2011iccv} or even the Harris corner detector~\citep{harris}. In the experimental section, we opt to use SIFT as the standard keypoint detector following the results and discussion in~\citep{jin2021image}, where SIFT keypoints provided competitive results for the downstream tasks of structure-from-motion and wide-baseline pairwise camera pose estimation.

\subsection{Geodesic Mapping Function}\label{sec:mapping_geo}

To obtain invariance to isometric non-rigid deformations in image space, we need a mapping function $f(.)$ that relates a physical point on the 3D surface to its correct projected pixel onto the image. Therefore, we propose to use geodesic isocurves defined in the manifold, to obtain invariance to isometric deformations of the surface. Once the geodesic curves around a keypoint are computed in the manifold as depicted in Figure~\ref{fig:teaserS-geodesic-aware}, one can sample 3D points from the surface that are geodesically equidistant from the center of the keypoins. This sampling strategy is by construction invariant to isometric deformations. After sampling points in the surface, we proceed to use the pinhole camera model to reproject the sampled 3D points to the image, attaining invariance to isometric deformations in image space.
	
	One possible approach to compute geodesic isocurves in the manifold is to use the Heatflow strategy~\citep{Crane:2013:TOG, nascimento2019iccv}. However, in this paper, we investigate a more efficient strategy for geodesics estimation on local keypoints. By considering a local estimation of the geodesic distances, we can efficiently construct a geodesic-aware image patch by remapping points on the surface onto the image with a geodesic walking strategy.

\begin{figure}[t!]
	\centering
	\includegraphics[width=0.9\linewidth]{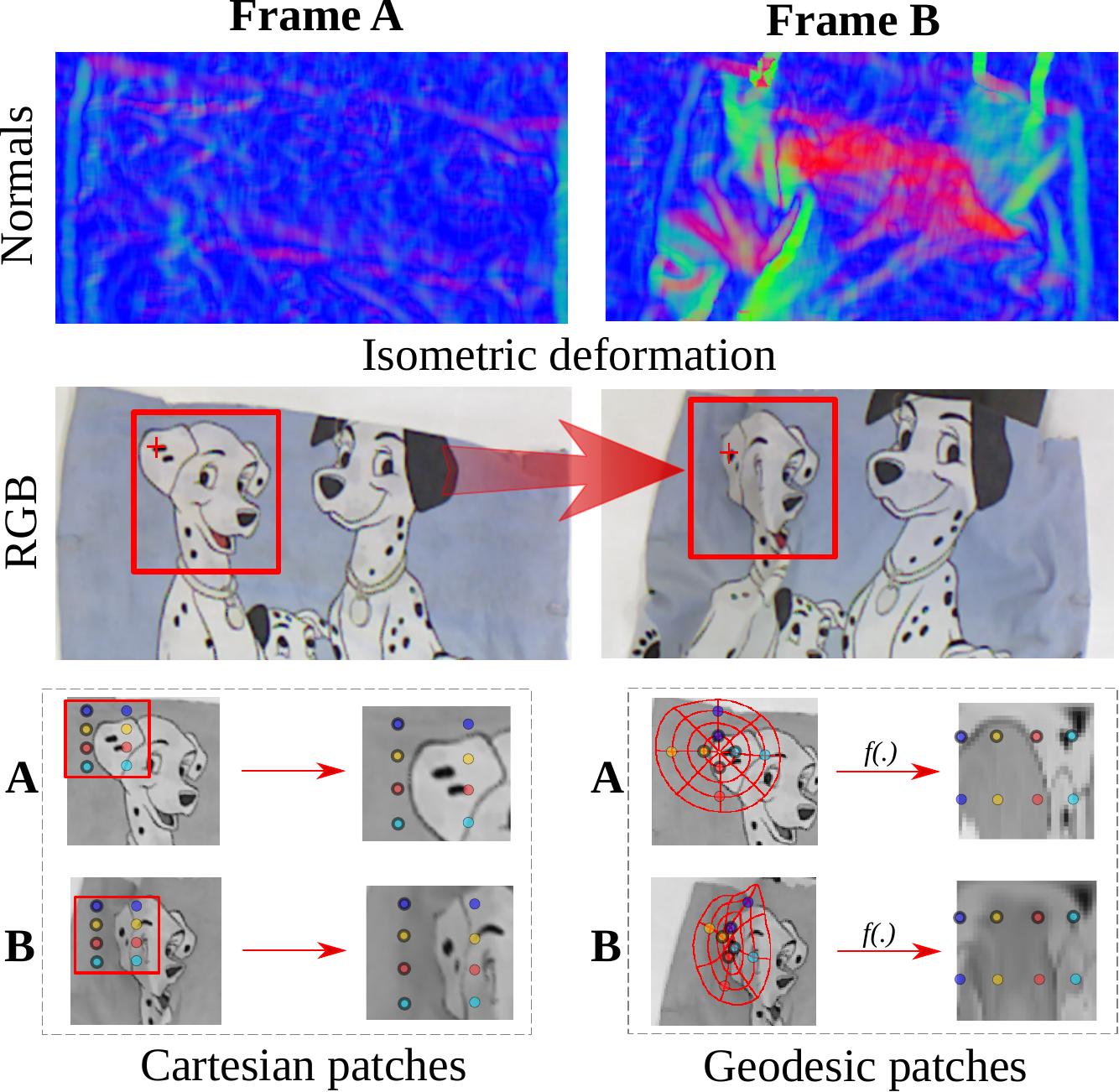}
	\caption{{\bf RGB-D patch rectified by the geodesic mapping function $f$}. The deformation is represented by the color changes in the surface normals from the depth (normal vectors orientations are encoded by color). The image intensity is sampled from a geodesic polar grid extracted from the mesh. The resulting intensities and gradients of the sampled geodesic patch maintains features of the original texture.
	}\label{fig:geodesic_mapping_function}
\end{figure}

\begin{figure}[t!]
	\centering 
	\includegraphics[width=0.88\columnwidth]{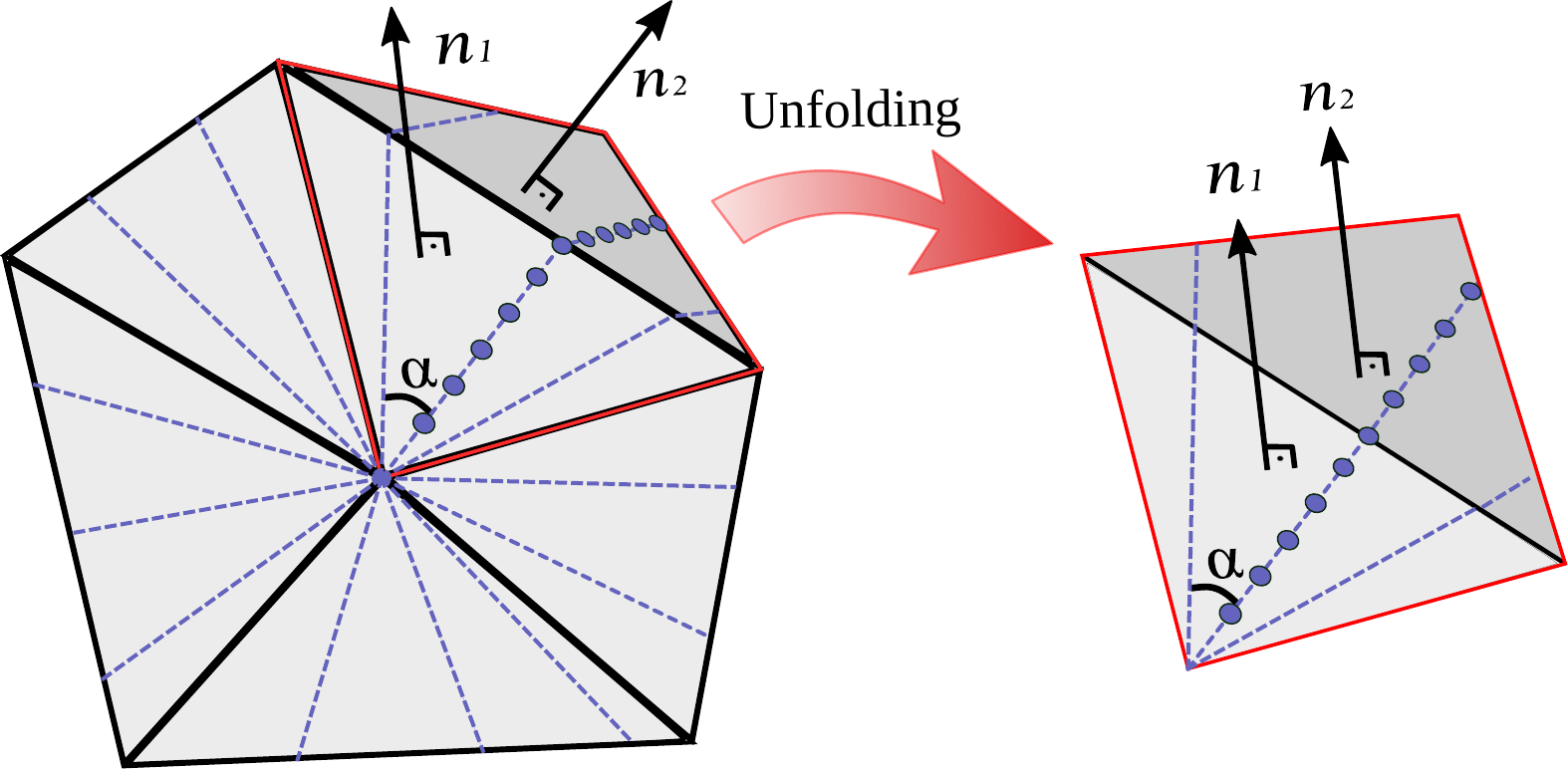}
	\caption{{\bf Geodesic paths computation in a local polar coordinate system}. Each walking direction is cast from the center vertex (keypoint). The angle $\alpha$ between each ray is defined by discretizing the unit circle by the number of desired angular bins (we used $32$ bins in our implementation). The blue dots along a path represent the points $(\theta_i, r_j), i \in \{1,...,m\}, j \in \{1,...,n\}$ that are equi-sampled in the $i$-th geodesic path.\label{fig:geopatch}}
\end{figure}

Given a mesh from a \textit{manifold} $\mM = (\mV, \mE)$, where  ${\mE \in \mV\times \mV}$ are the faces edges and ${\mV = \{\bv{v}_1, \dots, \bv{v}_k \}}$ the vertices $\in \mathbb{R}^3$, similar to the ISC descriptor approach~\citep{Kokkinos2012cvpr}, we define a local polar coordinate system $(\theta, r)$ on each keypoint $\bv k_k \in \mK$, where the center is the respective vertex $\bv{v}_k$ and $\theta$ is the angular coordinate and $r$ is the radial coordinate in geodesic units. Then, we construct a mapping function $f: \mathbb{R}^3 \to \mathbb{R}^2$ for an image patch $\mathbf{P}$ of resolution $m \times n$ that maps points on the manifold embedded in $ \mathbb{R}^3$ using the estimated geodesic distances, to the Cartesian coordinate system of an image patch (see \figurename~\ref{fig:geopatch}).
Since there is a rotation ambiguity, we arbitrarily set the canonical orientation to be axis-aligned with the RGB image; thus, the constructed patch is not invariant to image in-plane rotations yet, which will be handled in later steps. To sample the points in this coordinate system, the walking direction axis $\theta$ is discretized in $m$ uniform bins such as $\theta_i = 2\pi i/m$, $i \in \{1,\ldots,m\}$. For the distance coordinate $r$, a constant $\sigma$ (walking distance step) is chosen in order to sample the distance in $r_j = j \sigma$, $j \in \{1,\ldots,n\}$. 
Therefore, we construct the image patch by mapping the geodesic polar coordinates to the rectified patch $\mathbf{P}^{m \times n}$ by sampling $m$ orientations and, for each orientation, $n$ points in a geodesic path direction (the blue dots showed in \figurename~\ref{fig:geopatch}), which are geodesically equidistant from each other. 

The geodesic path is incrementally created by applying a similar idea to the Fast Marching algorithm~\citep{fastmarching}. Since our problem only requires the ray to be cast in a specific known direction on the manifold, we only need to find edge intersections and perform vector rotations in one direction, leading to a number of computations proportional to the size of the support region. This procedure is simpler and more efficient than Fast Marching.
Thus, let $\bv{v}_k$ be the coordinate system center. Each 3D ray vector $\mathbf{u}_i$ constructed from its respective angular bin $\theta_i$ is shot outward the center, and a geodesic path is created considering the initial direction of each bin. The initial directions lie on the plane of their respective faces, bounded by one-ring distant faces. Then, we compute the intersection of the direction and the next triangle edge. In order to continue the path, we rotate the direction $\mathbf{u}_i$ of the ray assuring that whenever the next face is unfolded to have the same normal as the current face, the result is a straight line between the triangles. 
%
%
We call this step \textit{unfolding} and it is illustrated in \figurename~\ref{fig:geopatch}. Each sampled 3D point $\mathbf{p}_{(.)}$ along the geodesic path is projected onto the RGB image coordinates using the perspective projection function $\pi$. 

\begin{algorithm}[tb!]
	
	\caption{\new{Geodesic Mapping Function $f(\cdot)$}}
	\SetKwInOut{Input}{inputs}
	\SetKwInOut{Output}{output}
	\SetKwProg{BuildPatch}{BuildPatch}{}{}
	
	\SetKwFunction{sampledirs}{sampleDirs}
	\SetKwFunction{intersect}{intersect}
	\SetKwFunction{unfold}{unfold}
	\SetKwFunction{sampleEquidistant}{sampleEquidistant}
	\SetKwFunction{reproject}{reproject2D}
	
	\BuildPatch{($\mM, \mI, \bv{v}_k, m, n, \sigma, \pi$)}{
		\Input{Mesh $\mM$; image $\mI$; keypoint vertex $\bv{v}_k$; the number of angular and radial bins $m, n$; sampling distance $\sigma$ and perspective projection function $\pi$.}
		\Output{The constructed $\mathbf{P}^{m \times n}$ patch.}
		$\mathbf{U} \gets \sampledirs(\mM, \bv{v}_k, m)$\;
		\ForEach{Direction $\mathbf{u}_i \in \mathbf{U}$}
		{
			$r_{max} \gets 0$\;
			$\mathcal{C} \gets \{\bv{v}_k\}$ \tcc*[r]{Intersection set}
			$\mathbf{c}_{current}\gets \bv{v}_k$\;
			\While{$r_{max} < n\sigma$}
			{
				$\mathbf{c}_{next} \gets \intersect(\mM,\mathbf{c}_{current},\mathbf{u}_i)$\;
				$\mathbf{u}_i \gets \unfold(\mM,\mathbf{c}_{next},\mathbf{u}_i)$\;
				$\mathcal{C} \gets \mathcal{C} \cup \{\mathbf{c}_{next}\}$\;
				$r_{max}\mathrel{+}= ||\mathbf{c}_{next} - \mathbf{c}_{current} ||_2$\;
				$\mathbf{c}_{current} \gets \mathbf{c}_{next}$ \;
			}
			
			$\mathcal{G} \gets \{\}$ \tcc*[r]{Geodesic paths}
			$\mathcal{G} \gets \sampleEquidistant(\mathcal{C}, n, \sigma)$\;

			\ForEach{Sampled 3D point $\mathbf{p}_j \in \mathcal{G}$}
			{
				$\mathbf{\tilde{p}}_j \sim \pi(\mathbf{p}_j)$ \;
				$\mathbf{P}_{[i, j]} \gets \mI(\mathbf{\tilde{p}}_j)$\;		
			}
			
		}
		\KwRet{$\mathbf{P}$}\;
	}  \label{alg:geopatch}
\end{algorithm}

The Algorithm~\ref{alg:geopatch} shows the steps to construct the geodesic image patch. The perspective projection function $\pi$ returns the pixel coordinates $\tilde{\mathbf{p}}_{(.)}$ in image space that are used to obtain the scalar values from the intensity image. The intensity values are obtained via bilinear interpolation on the unit square:
	
	\begin{equation}
	\mI(\tilde{\mathbf{p}}_{(.)}) = \begin{bmatrix} 1 - x & x \end{bmatrix} 
	\begin{bmatrix}
	\tilde{\mI}(0,0) & \tilde{\mI}(0,1) \\
	\tilde{\mI}(1,0) & \tilde{\mI}(1,1)
	\end{bmatrix}
	\begin{bmatrix} 1 - y \\ y \end{bmatrix},
	\end{equation}
	
	\noindent where $x$ and $y$ are the decimal parts of the projected point $\tilde{\mathbf{p}}_{(.)}$ and $\tilde{\mI}$ gives the known intensity values in the image $\mI$ corrected by the offset that brings the integer coordinates of $\tilde{\mathbf{p}}_{(.)}$ to the origin.
	
	Given the current triangle of the walking step $\mathbf{t}_{current} = (\mathbf{v}_a,\mathbf{v}_b,\mathbf{v}_c)$, and the triangle normal $\vec{\mathbf{n}}$, all obtained using the mesh data and current walking intersection point $\mathbf{c}_{current}$,  the \textit{intersect} routine calculates the next intersection point of an edge, given the walking direction with the line-plane intersection:
	\begin{equation}
	\mathbf{c}_{next} = \mathbf{c}_{current} + d  \mathbf{u}_i, \mbox{ and }   d = \dfrac{(\mathbf{v}_{out} - \mathbf{c}_{current}) \cdot \mathbf{n}_p}{\mathbf{u}_i \cdot \mathbf{n}_p}.
	\end{equation}
	\noindent $d$ is the positive solution to the line-plane intersection, $\mathbf{n}_p = \mathbf{n} \cross \mathbf{e}_{out}$ is the normal of the plane defined by the triangle normal $\mathbf{n}$ and escape edge $\mathbf{e}_{out}$ (the edge that first intersects with $\mathbf{u}_i$), and finally $\mathbf{v}_{out}$ is a vertex from the escape edge $\mathbf{e}_{out}$. The \textit{unfold} function rotates $\mathbf{u}_i$, with the Rodrigue's formula, to the next face when it crosses an edge:
	\begin{equation}
	\mathbf{u}_i = \mathbf{u}_i \cos{\varphi} + (\mathbf{k} \cross{\mathbf{u}_i}) \sin{\varphi} + \mathbf{k} (\mathbf{k}\cdot{\mathbf{u}_i) (1 - \cos{\varphi})},
	\end{equation}
	
	\noindent where $\mathbf{k} = \mathbf{n}_1 \cross{\mathbf{n}_2}$ is the orthogonal direction, and $\varphi$ is the angle between the current and next faces' normals $\mathbf{n}_1$ and $\mathbf{n}_2$, respectively. The \textit{unfold} step is illustrated in Figure~\ref{fig:geopatch}, and is repeated until the maximum geodesic distance is met.

Figure~\ref{fig:geodesic_mapping_function} shows a real deformation example, where the geodesic patch is compared to the classic Cartesian sampling approach. One can see that the rectified patch using a classic Cartesian sampling presents textures such as the eyes of the dog, that were not present in the original undeformed patch. Conversely, the geodesic patch kept the features accordingly to the undeformed template.

\subsection{Geodesic-Aware Feature Extraction} \label{sec:geopatch}

After mapping pixels locations from a deformable surface to a rectified image patch $\mathbf{P}$ using the mapping function $f$, we proceed to extract the visual features $\mathbf{d}_f$. The key idea in this strategy is mapping each pixel's location, preserving the intrinsic geometry of the surface, and further analyzing the vicinity of the keypoint. We present two novel efficient geodesic-aware descriptors. The first, called {\it GeoBit}, is a binary descriptor that encodes deformation-invariant features using a vector of binary tests and the second method, named {\it GeoPatch}, is based on learning the feature extraction with a shallow convolutional neural network. Since both GeoBit and GeoPatch encode pixel locations using the geodesic distance, the visual pattern remains invariant in the presence of isometric deformations of the surface.

\paragraph{Binary feature extraction (GeoBit)} The GeoBit descriptor exploits visual and geometrical information to encode deformation-invariant features into a binary vector. 

\begin{figure}[t!]
	\centering
	\includegraphics[width=0.7\linewidth]{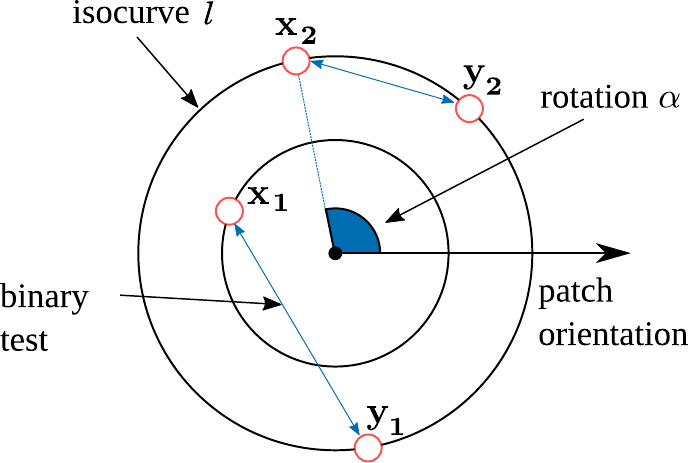}
	\caption{\new{{\bf Two binary tests to extract the visual features}. For each binary test in the pattern, we store the isocurve $l$ and the rotation $\alpha$ \wrt the patch orientation of two points.}}
	\label{fig:pairselectionexample}
\end{figure}

After computing the geodesic patch of the keypoint's neighborhood, we compute the visual features based on a predefined set of binary intensity tests over the polar coordinates. Thus, we select a set of pairs of pixels from patch $\mathbf{P}$ to create a gradient field to extract the visual pattern. 
Given an image keypoint $\bv k \in \mK$, we extract the rectified geodesic patch $\bf P$ centered at $\bv k$ as discussed in Section \ref{sec:mapping_geo}. We then sample pixel pairs around the keypoint $\bv k$ using a fixed pattern with locations given by a distribution. We store, for each point in the pattern, the isocurve $l$ at the point, and the rotation angle $\alpha$ \wrt the patch orientation, as illustrated in \figurename~\ref{fig:pairselectionexample} with two test pairs of points lying onto the isocurves. We can then build the set $\mathcal{S}=\{\left( \bv{x}_i, \bv{y}_i \right), i=1,\dots,n\}$, as the fixed set of sampled pairs from $\bv{P}$, where  $\bv{x}_i$ and $\bv{y}_i$ encode the isocurve $l$ and angle $\alpha$ of the $i$-th pixel of the binary test pair, \ie, $\bv{x_i} = (\alpha_i, l_i)^T$. The extracted descriptor from the patch $\bv{P}$ associated with the keypoint $\bv{k}$ is then represented as the binary string: 
\begin{equation}
\mathbf{d}_f = \sum_{1}^{n} 2^{i-1} [\bv{P}(\bv{x}_i) < \bv{P}(\bv{y}_i)],
\end{equation}
\noindent where $\bv{P}(\bv{x}_i)$ returns the pixel intensity in the polar coordinates $\bv{x}_i$ and $[t]$ is the Iverson bracket that returns $1$ if the predicate $t$ is true and $0$ otherwise. The comparison in the bracket captures gradient changes in the keypoint neighborhood. 

Similar to DaLI descriptor, in GeoBit, the invariance to rotation is achieved by rotating the tests' set in polar coordinates ${\mathcal{S}_{\theta} = \{(\bv{T}_{\theta}(\bv{x}_i), \bv{T}_{\theta}(\bv{y}_i)) | (\bv{x}_i,\bv{y}_i) \in \mathcal{S}\}
	}$ after transforming all tests $\mathcal{S}$ in the first axis $\theta$ by a horizontal shift $\bv{T}_{\theta}$.

Therefore for each keypoint we compute a set of candidate descriptors (we used $16$ candidates in our experiments) with different orientations by rotating the coordinates of the pattern points in set $\mathcal{S}$ using discretized rotations uniformly sampled from $[0,2\pi]$, \ie, adding $\theta_n = n\pi/8, ~n \in \{0,\ldots,15\}$ to the first coordinate $\mathbf{T}_{\theta_n}(\mathbf{x}_i) = (\alpha_i + \theta_n, l_i)$.

In the matching step, we select the feature vector with an orientation that results in the smallest hamming distance between two compared keypoints. This strategy has shown better performance when compared to calculating the canonical orientation for each keypoint using gradient-based approaches, mainly because non-rigid deformations around the keypoints introduce additional noise in the orientation estimation.

\new{
The Geobit implementation has a dimension of $512$ bits for each computed orientation, and $16$ possible orientations, resulting in a memory footprint of $1{,}024$ bytes.
The direction changes of gradients around the keypoint are computed using image intensity comparison tests, which have small memory storage requirements and can be matched very efficiently in modern CPUs by leveraging vectorized low-level instructions.
}

%
%


\paragraph{Geodesic-Guided Feature Learning (GeoPatch)} Despite sharing the mapping function for localizing pixels, GeoPatch differs from GeoBit in how the visual features are extracted from the resulting sampled pixel intensities. While GeoBit is based on a fixed pattern that indicates which pixels will be evaluated by pairwise intensity tests, GeoPatch uses a convolutional neural network approach, which has achieved state-of-the-art performance in feature extraction, on the resulting sampled intensities mapped to a regular grid, enabling efficient extraction of distinctive and invariant patterns.

\begin{figure}[!t]
	\centering
	
	\includegraphics[width=0.86\linewidth]{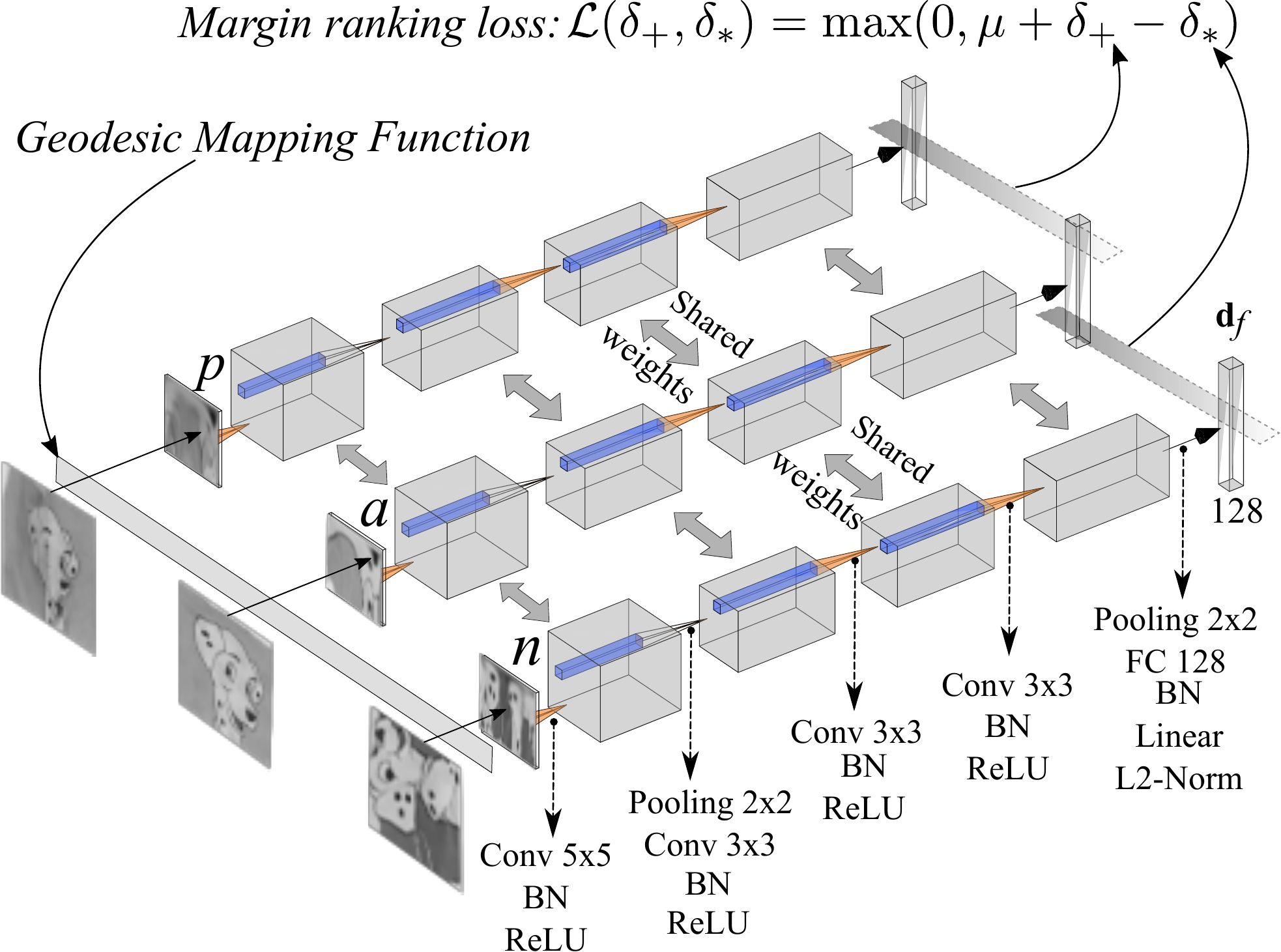}
	\caption{{\bf Triple siamese architecture of GeoPatch descriptor}. During the training, we fed a siamese ConvNet with the patch triplets $(\mathbf{a}, \mathbf{p}, \mathbf{n})$. The margin ranking loss is computed  and the errors are back-propagated. After the training stage, one branch is used to extract a $128$-dimensional descriptor $\mathbf{d}_f$.}\label{fig:architecture_siamese}

\end{figure}

A $128$-dimensional floating-point feature vector $\mathbf{d}_f$ for an image patch $\mathbf{P}$ is computed by forward-propagation of the rectified patch $\mathbf{P}$ in the network $\mathbf{G}$, \ie, $\mathbf{d}_f = \mathbf{G(P)} $. We adopted a variation of a shallow convolutional architecture from~\cite{tfeat-balntas2016} to build $\mathbf{G}$. \figurename~\ref{fig:architecture_siamese} shows the geodesic patch estimation and the network architecture. Although in this work, we chose a compact network that provides performance and generalization, deeper and more complex networks can be employed as well. GeoPatch demonstrates that we can efficiently use a classic ConvNet to handle a non-Euclidean geometric task. By using the proposed geodesic image patch to feed a compact ConvNet, we do not overwhelm the network to handle many different image transformations, which inadvertently leads to overfitting. Moreover, we take advantage of the network's powerful ability to learn feature maps that can be used to extract a compact but distinctive descriptor.

The network $\mathbf{G}$ is trained using $N$ triplets of patches ${(\mathbf{a}^{(i)}, ~\mathbf{p}^{(i)}, \mathbf{n}^{(i)})}$, where $\mathbf{a}$ is an anchor patch, $\mathbf{p}$ is a positive corresponding patch (a projection of the same keypoint in different views), and $\mathbf{n}$ is a non-matching (negative) patch of the keypoint. The triplets are built by sampling indices of patches from a uniform distribution, considering a training dataset with annotated image matches. We use the margin ranking loss function with an anchor swap to train the network. The margin ranking loss with anchor swap considers the $L_2$ distance $\delta$ between the computed descriptors after a mini-batch forward pass:
\begin{equation}
\mathcal{L}\left(\delta_{+}^{(.)}, \delta_{*}^{(.)} \right) = \dfrac{1}{N} \sum_{i=1}^{N} \max(0, \mu + \delta_{+}^{(i)} - \delta_{*}^{(i)}),
\end{equation}

\noindent where ${\delta_{*} = \min(\left\Vert \bv{G}(p) - \bv{G}(n) \right\Vert_2, \left\Vert \bv{G}(a) - \bv{G}(n) \right\Vert_2)}$  is the hardest negative distance in the triplet, since there is two possible negative distances (while there is only one possible positive distance), and $\delta_{+} =  \left\Vert \bv{G}(p) - \bv{G}(a) \right\Vert_2$ is the distance between the positive and anchor patches. Finally, $\mu$ is a positive scalar encoding the margin's length. By using the hardest negative distance in the triplet, the gradient updates are larger, which means that the network is learning more with little computation overhead since no extra forward passes are needed, and the three possible distance computations in the triplet are negligible. When minimizing the margin ranking loss, the network seeks to bring closer matching patches in the descriptor space, while it tries to push out non-matching ones of the hypersphere defined by the margin $\mu$.
The triple siamese network architecture used during training is detailed in \figurename~\ref{fig:architecture_siamese}.

Our network implementation employs circular padding in the horizontal axis before all convolution layers to take advantage of the circularity of the polar grid sampling in the patch construction. Due to the equivariance property of the polar sampling, rotations in the polar space are converted to horizontal shifts in the sampled patches. In this context, the max-pooling operations provide invariance to small translations in the patches, \ie, to rotations of up to $4$ pixels in the image, since there are two $2 \times 2$ max pooling operations.
To achieve invariance to rotation, we perform max pooling in the horizontal axis of the tensor (angle axis in geodesic patch) before  the fully-connected layer. We also tested to augment the data by applying horizontal shifts to the patches during training, however, the first method works slighly better according to our experiments.

In the last layer of the network, the convolutional filters are flattened and forwarded to a fully connected layer with linear activation function, projecting the features to a $128$ dimensional space, which are then $L_2$ normalized to produce the final descriptor feature vector $\bv{d}_f$.





%% file: experiments.tex
\section{Experiments}\label{sec:exp}


We evaluate GeoBit and GeoPatch with both simulated and real data and compare the results against different descriptors. The simulated datasets provide a tractable but realistic set up to test specific behavior of the descriptors, while the real data demonstrate the applicability of the approach on real scenarios.

\subsection{RGB-D Non-Rigid Datasets}\label{sec:dataset}

In this section, we introduce a benchmark composed of two datasets of real-world objects acquired with two different RGB-D sensors, and an additional simulated dataset with thousands of RGB-D images of deforming objects.

\begin{figure}[t!]
	\centering
	\includegraphics[width=0.9\columnwidth]{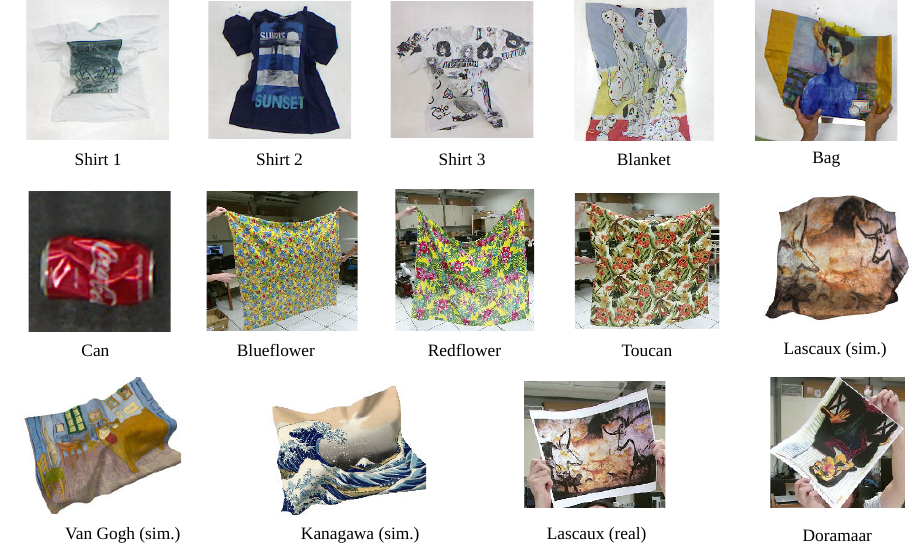}
	\caption{{\bf Examples of real-world and simulated (sim.) data in our dataset}. The objects are subjected to deformations while frames are acquired with an RGB-D camera. For each object, we provide ground-truth keypoint correspondences that can be used to evaluate the performance of local descriptors.}
	\label{fig:dataset}
\end{figure}

\paragraph{Real-World Data} To evaluate the matching capability of our descriptor on real-world images, we built a new dataset of deforming objects~\footnote{Publicly available at \url{https://www.verlab.dcc.ufmg.br/descriptors}} comprising $11$ deformable objects, which are split into two sets considering the sensor used and the annotation method, as shown in \figurename~\ref{fig:dataset}. The RGB-D images were captured with Kinect{\texttrademark}, versions 1 and 2. In the sequences acquired with Kinect 1, the images have resolution of $640 \times 480$ pixels, and image correspondences of a set of landmarks were manually annotated. In the Kinect 2 sequences, the images were acquired at resolution $1{,}920 \times 1{,}080$ with and flat reflective markers tightly fixed behind the objects' surfaces, determining the position of the landmarks. The correspondences were obtained with a high precision motion capture system (OptiTrack\texttrademark). 

We developed a refined deformation model between each image pair that can be used to establish dense correspondences of detected keypoints reliably. For this purpose, it is employed a thin-plate-spline (TPS)~\citep{tps} deformation model between image frames. Since the TPS model, in general, tends to have larger errors in regions far from the control points, a coarse deformation model is estimated through the sparse set of annotated correspondences, \ie, the landmarks' location, which will be close to global optima with respect to the Structure Similarity (SSIM) metric. The control points are densified by regularly sampling points in the source image and the deformation model parameters are then photometrically optimized via gradient descent, resulting in a refined warping model. 

We also performed experiments with the public dataset DeSurT~\citep{wang2019deformable}, which contains challenging RGB-D sequences of deforming objects. Several objects from DeSurT exhibit textureless and repetitive textures, posing challenges to all local descriptors, complementing our datasets in these aspects. The pixels with annotated correspondences are used as landmarks to compute the TPS model for the DeSurT dataset.

\paragraph{Synthetic Data} \label{sec:synth} We used a physics simulation of cloth to create arbitrary non-rigid isometric deformations with ground-truth correspondences. Considering a grid of particles having mass and a 3D position, Newton's second law is applied in conjunction with Verlet integration, to act over the particles' position. 
A constraint satisfaction optimization step is performed over all particles to enforce constant distance of neighboring particles, thus keeping the deformation isometric. The details of the physics simulation are described in the Supplementary material.

The texture is applied onto the mesh generated by the grid and rendered with diffuse illumination as the cloth moves (which causes non-linear illumination changes). We project the grid of particles onto the image and use the Harris corner score to retain approximately the $100$ best corner-like features. These $100$ points are used as landmarks to compute the TPS model. 
For each simulation, we use random inputs for the simulation parameters, including wind force, wind direction, illumination strength, illumination position, and Gaussian noise in the image pixels, ensuring different outcomes in every run, resulting in realistic images and rich deformations. To obtain plausible textures for the simulations, we built a broad set of images by merging several Structure-from-Motion image collections~\citep{cit:1dsfm}. 
We ran the simulation with different parameters and grabbed $30$ snapshots, generating a set of $30$ RGB-D images with ground-truth correspondences for each input texture.

\begin{table*}[t!]

	\centering
	\caption{{\bf Comparison with state of the art}. Our descriptors are able to provide higher matching scores in the sequences from our dataset and the DeSurT dataset. Best in bold, second-best in italic.}
	\label{table:matching_score}
	
	\resizebox{0.95\linewidth}{!}{%
		
		\begin{tabular}{@{}clccccccccc@{}}
			\toprule 
			\multirow{3}{*}{\bf Dataset} & \multirow{3}{*}{\bf Object (\# pairs)} & \multicolumn{9}{c}{{\bf Avg. Matching Scores}}  \\ \cmidrule{3-11} 
			& \multicolumn{1}{r}{} & \rotatebox[origin=c]{0}{\parbox[c]{1.5cm}{\centering BRAND}}& \rotatebox[origin=c]{0}{\parbox[c]{1.5cm}{\centering DAISY}}& \rotatebox[origin=c]{0}{\parbox[c]{1.5cm}{\centering DaLI}}& \rotatebox[origin=c]{0}{\parbox[c]{1.5cm}{\centering FREAK}}& \rotatebox[origin=c]{0}{\parbox[c]{1.5cm}{\centering Log-Polar}}& \rotatebox[origin=c]{0}{\parbox[c]{1.5cm}{\centering ORB}}&  \rotatebox[origin=c]{0}{\parbox[c]{1.5cm}{\centering TFeat}}& \rotatebox[origin=c]{0}{\parbox[c]{1.5cm}{\centering GeoBit}}& \rotatebox[origin=c]{0}{\parbox[c]{1.5cm}{\centering GeoPatch}} \\
			
			\multirow{6}{*}{\rotatebox[origin=c]{90}{\parbox[c]{1.5cm}{\centering Kinect 1}}} 
			
			& \multicolumn{1}{l}{Shirt2 ($18$)}& $0.24$ & $0.30$ & $0.35$ & $0.33$ & $0.36$ & $0.25$ & $0.32$ & $\mathbf{0.44}$ & $\mathit{0.43}$  \\
			& \multicolumn{1}{l}{Blanket1 ($15$)}& $0.15$ & $0.25$ & $0.28$ & $0.22$ & $0.29$ & $0.20$ & $0.26$ & $\mathbf{0.34}$ & $\mathit{0.33}$  \\
			& \multicolumn{1}{l}{Shirt3 ($17$)}& $0.21$ & $0.27$ & $0.28$ & $0.28$ & $0.30$ & $0.22$ & $0.27$ & $\mathbf{0.35}$ & $\mathit{0.34}$  \\
			& \multicolumn{1}{l}{Can1 ($6$)}& $0.07$ & $0.05$ & $0.09$ & $0.07$ & $\mathit{0.16}$ & $0.05$ & $0.13$ & $\mathbf{0.18}$ & $\mathit{0.16}$  \\
			& \multicolumn{1}{l}{Bag1 ($4$)}& $0.12$ & $0.08$ & $0.15$ & $0.18$ & $0.22$ & $0.10$ & $0.16$ & $\mathbf{0.30}$ & $\mathit{0.28}$  \\
			& \multicolumn{1}{l}{Shirt1 ($14$)}& $0.17$ & $0.26$ & $0.25$ & $0.26$ & $0.30$ & $0.20$ & $0.26$ & $\mathbf{0.34}$ & $\mathbf{0.34}$  \\
			
			\midrule
			\multirow{11}{*}{\rotatebox[origin=c]{90}{\parbox[c]{1.5cm}{\centering Kinect 2}}}
			& \multicolumn{1}{l}{doramaar\_l ($29$)}& $0.28$ & $0.35$ & $\mathit{0.41}$ & $\mathit{0.41}$ & $0.35$ & $0.34$ & $0.35$ & $0.40$ & $\mathbf{0.42}$  \\
			& \multicolumn{1}{l}{toucan\_m ($29$)}& $0.26$ & $0.33$ & $\mathbf{0.38}$ & $\mathit{0.36}$ & $0.32$ & $0.28$ & $0.30$ & $0.31$ & $0.35$  \\
			& \multicolumn{1}{l}{blueflower\_l ($29$)}& $0.27$ & $0.29$ & $0.37$ & $0.36$ & $0.30$ & $0.24$ & $0.27$ & $\mathit{0.38}$ & $\mathbf{0.39}$  \\
			& \multicolumn{1}{l}{blueflower\_h ($29$)}& $0.20$ & $0.28$ & $\mathbf{0.37}$ & $0.30$ & $0.30$ & $0.24$ & $0.27$ & $\mathit{0.36}$ & $\mathit{0.36}$  \\
			& \multicolumn{1}{l}{toucan\_h ($29$)}& $0.24$ & $0.31$ & $\mathbf{0.35}$ & $\mathit{0.34}$ & $0.33$ & $0.26$ & $0.29$ & $0.28$ & $0.33$  \\
			& \multicolumn{1}{l}{redflower\_h ($29$)}& $0.12$ & $0.12$ & $\mathit{0.17}$ & $0.16$ & $0.13$ & $0.11$ & $0.12$ & $\mathbf{0.19}$ & $\mathbf{0.19}$  \\
			& \multicolumn{1}{l}{blueflower\_m ($29$)}& $0.19$ & $0.27$ & $0.35$ & $0.30$ & $0.30$ & $0.23$ & $0.26$ & $\mathit{0.36}$ & $\mathbf{0.37}$  \\
			& \multicolumn{1}{l}{toucan\_l ($29$)}& $0.26$ & $0.35$ & $\mathit{0.37}$ & $\mathbf{0.38}$ & $0.36$ & $0.29$ & $0.34$ & $0.29$ & $0.35$  \\
			& \multicolumn{1}{l}{lascaux\_l ($29$)}& $0.33$ & $0.46$ & $0.52$ & $0.53$ & $0.48$ & $0.44$ & $0.47$ & $\mathit{0.55}$ & $\mathbf{0.56}$  \\
			& \multicolumn{1}{l}{redflower\_m ($29$)}& $0.14$ & $0.18$ & $0.24$ & $0.21$ & $0.19$ & $0.15$ & $0.18$ & $\mathit{0.27}$ & $\mathbf{0.28}$  \\
			& \multicolumn{1}{l}{redflower\_l ($29$)}& $0.18$ & $0.21$ & $0.29$ & $0.25$ & $0.22$ & $0.18$ & $0.20$ & $\mathit{0.30}$ & $\mathbf{0.31}$  \\

			\midrule
			\multirow{11}{*}{\rotatebox[origin=c]{90}{\parbox[c]{1.5cm}{\centering DeSurT}}} 
			& \multicolumn{1}{l}{brick ($29$)}& $0.20$ & $0.16$ & $\mathit{0.31}$ & $0.22$ & $0.28$ & $0.25$ & $0.26$ & $\mathit{0.31}$ & $\mathbf{0.32}$  \\
			& \multicolumn{1}{l}{campus ($29$)}& $0.16$ & $0.22$ & $\mathit{0.30}$ & $0.17$ & $0.28$ & $0.21$ & $0.25$ & $0.28$ & $\mathbf{0.32}$  \\
			& \multicolumn{1}{l}{sunset ($29$)}& $0.08$ & $0.12$ & $0.11$ & $\mathit{0.13}$ & $\mathbf{0.15}$ & $0.09$ & $0.12$ & $0.12$ & $\mathbf{0.15}$  \\
			& \multicolumn{1}{l}{cushion1 ($29$)}& $0.11$ & $0.17$ & $0.15$ & $0.14$ & $0.20$ & $0.13$ & $0.18$ & $\mathit{0.21}$ & $\mathbf{0.23}$  \\
			& \multicolumn{1}{l}{scene ($29$)}& $0.21$ & $0.19$ & $\mathit{0.29}$ & $0.20$ & $0.28$ & $0.23$ & $0.27$ & $\mathit{0.29}$ & $\mathbf{0.32}$  \\
			& \multicolumn{1}{l}{cushion2 ($29$)}& $0.07$ & $0.06$ & $0.07$ & $0.08$ & $\mathit{0.09}$ & $0.06$ & $0.07$ & $\mathbf{0.11}$ & $\mathit{0.09}$  \\
			& \multicolumn{1}{l}{cobble ($29$)}& $0.16$ & $0.09$ & $0.21$ & $0.15$ & $0.24$ & $0.21$ & $0.22$ & $\mathit{0.26}$ & $\mathbf{0.28}$  \\
			& \multicolumn{1}{l}{newspaper2 ($29$)}& $0.15$ & $0.20$ & $0.21$ & $0.18$ & $\mathit{0.24}$ & $0.17$ & $0.21$ & $\mathit{0.24}$ & $\mathbf{0.26}$  \\
			& \multicolumn{1}{l}{newspaper1 ($29$)}& $0.16$ & $0.23$ & $0.27$ & $0.19$ & $\mathit{0.32}$ & $0.24$ & $0.28$ & $\mathit{0.32}$ & $\mathbf{0.34}$  \\
			
			\midrule
			\multirow{7}{*}{\rotatebox[origin=c]{90}{\parbox[c]{1cm}{\centering Simulation}}}
			& \multicolumn{1}{l}{kanagawa\_rot ($18$)}& $0.06$ & $0.22$ & $0.13$ & $0.21$ & $0.12$ & $0.19$ & $0.27$ & $\mathit{0.31}$ & $\mathbf{0.35}$  \\
			& \multicolumn{1}{l}{lascaux\_rot ($18$)}& $0.08$ & $0.36$ & $0.26$ & $0.31$ & $0.18$ & $0.33$ & $0.38$ & $\mathit{0.44}$ & $\mathbf{0.47}$  \\
			& \multicolumn{1}{l}{kanagawa\_scale ($3$)}& $0.02$ & $0.11$ & $0.01$ & $0.06$ & $0.35$ & $0.04$ & $0.33$ & $\mathit{0.45}$ & $\mathbf{0.46}$  \\
			& \multicolumn{1}{l}{chambre\_scale ($3$)}& $0.02$ & $0.06$ & $0.01$ & $0.03$ & $0.17$ & $0.03$ & $0.16$ & $\mathit{0.23}$ & $\mathbf{0.24}$  \\
			& \multicolumn{1}{l}{chambre\_rot ($18$)}& $0.06$ & $0.25$ & $0.20$ & $0.18$ & $0.15$ & $0.20$ & $0.26$ & $\mathit{0.29}$ & $\mathbf{0.34}$  \\
			& \multicolumn{1}{l}{lascaux\_scale ($3$)}& $0.02$ & $0.10$ & $0.02$ & $0.08$ & $0.35$ & $0.06$ & $0.33$ & $\mathit{0.43}$ & $\mathbf{0.45}$  \\
			
			\midrule
			Mean & \multicolumn{1}{c}{$*$}& $0.15$ & $0.22$ & $0.24$ & $0.23$ & $0.26$ & $0.20$ & $0.25$ & $\mathit{0.31}$ & $\mathbf{0.33}$ \\
			
			\bottomrule 
		\end{tabular}
		
	}
	
\end{table*}

\subsection{Implementation Details} 
Similar to classic RGB descriptors, the main parameter of our descriptors is the size of the support region around a keypoint defined by the isocurve thickness. By testing a set of empirically chosen values for all descriptors on the Bag1 dataset, we experimentally found that using a support region of $75$ millimeters for the geodesic patch, and keypoint radius of $15$ pixels for the RGB descriptors, produce slightly better recognition rates for all methods. The support region size is fixed in all experiments. 

\paragraph*{Depth preprocessing} 
In our pipeline, we apply a preprocessing step that denoises the depth values and fill missing data when possible. Therefore, we implement a multi-resolution strategy. First, we sub-sample the depth employing a smoothing pyramid of depth two, as it demonstrated to have the best trade-off between denoising and resolution in our experiments. To fill the holes, we first segment the depth image and detect blob regions where there is missing data. Then, we use the Inverse Distance Weighting interpolation by considering the known depth values around the blobs to fill the missing data. When the blob perimeter is larger than $400$ pixels, we skip it since there are too many missing values to fill.
Finally, we connect the neighboring pixels in the depth image to construct a triangular mesh, which is very fast and provides a mesh of sufficient quality for the geodesic distance estimation steps.

\paragraph{Binary Tests Distribution for GeoBit} GeoBit performs binary tests in the neighborhood around the keypoint. These tests are based on a set of pixels selected by a distribution function. As demonstrated in~\cite{nascimento2019iccv}, by considering i) An isotropic Gaussian distribution $\mathcal{N}(0,\frac{30^2}{100})$; and ii) a uniform distribution, where they randomly selected $1{,}024$ different angles and isocurves, the Gaussian distribution yields a slight improvement in performance. Thus, we employ the Gaussian distribution with predefined parameters following ~\cite{nascimento2019iccv}.

\paragraph{Convolutional Network Training}
To train the CNN, we used Stochastic Gradient Descent (SGD) with learning rate $lr = 0.1$, weight decay of $10^{-4}$ on all weights including bias, and a batch size of $1{,}000$ samples. 
The network was entirely trained on simulation data generated by the described approach in Subsection~\ref{sec:synth}. The simulated dataset images used in the experiment comes from an independent simulation instance, using unrelated images as textures, \ie, there is no data from the synthetic data of our dataset in the training stage.


\subsection{Matching Performance} \label{sec:results}

Given two sets of keypoints detected with SIFT~\citep{lowe2004ijcv} from a reference and a target image respectively, the task of matching the keypoints consists in, for each keypoint in the reference set, find its correspondence in the other set in case the keypoint is visible in both sets.

We compared our results to well-known binary descriptors for 2D images: ORB~\citep{rublee2011iccv} and FREAK~\citep{cit:alahi2012freak}; a floating-point gradient based descriptor for 2D images: DAISY~\citep{daisy}; a descriptor that combines texture and shape from RGB-D images: BRAND~\citep{nascimento2012iros}; a deformation-invariant descriptor for RGB images: DaLI~\citep{simo2015dali}; and two learning methods that works on grayscale image patches, TFeat~\citep{tfeat-balntas2016} and Log-Polar~\citep{cit:beyondcartesian}, the latter being an improvement of HardNet~\citep{cit:hardnet}. For all methods, the input grayscale intensity image $\mI$ is obtained from RGB color with the formula $\mI = 0.299R + 0.587G + 0.114B$.

Our performance assessment was conducted using the matching score metric as defined by \cite{cit:matchscore} and SIFT keypoints detected independently for each frame, following the protocol proposed in~\cite{jin2021image}. The matching scores is capable of measuring the performance of the descriptors in realistic conditions, where the independently detected keypoints might not be the same between frames. After detecting SIFT keypoints, we selected the $2{,}048$ most salient keypoints according to the response attribute. The ground-truth matches are given by the estimated ground-truth deformation model between the two frames. In our experiments, we matched the keypoint descriptors from pairs of images using brute force nearest neighbor search, \ie, for each descriptor in the reference set, we compare it with all descriptors in the target set and consider the descriptor with the smallest Euclidean distance (the nearest neighbor) as its corresponding descriptor. We labeled valid matches with two keypoints corresponding to the same physical location (according to the ground-truth) as positive, and as negative otherwise. The matching score metric is given by the number of correct matches, obtained using the refined deformation model, divided by the smallest number of keypoints detected in both images for an image pair. It is noteworthy that correspondence with detected keypoints is harder than using the annotated ones since several keypoints might not have true correspondences.

Table~\ref{table:matching_score} shows the matching scores for all descriptors in our experiments. It is worth mentioning that the rotation attributes of SIFT keypoints are shared among all descriptors. 
We note that among all methodologies, our descriptor GeoPatch stands out as the descriptor with the highest averages in matching score over different deformations, followed by GeoBit, in second place. For instance, the GeoBit descriptor is able to achieve slightly better matching scores on Kinect 1 dataset due to increased sensor noise. In this specific scenario, the Gaussian sampling used by GeoBit provides increased robustness against depth noise, since the spatial distribution of the tests are closer to the keypoint's center.
From these results, we can draw the following additional observations. First, BRAND performance is drastically reduced by deformations since it is based on the normals of a support region, which is not an intrinsic property of a surface, hence not being invariant to non-rigid isometric deformations. Second, the photometric information is also impaired by the deformations, which penalizes twice RGB-D descriptors not aware of deformations like BRAND. 

The performance of the RGB descriptors is also directly impacted by the deformations, since all of them use a sampling strategy over image pixels defined in the Cartesian space, and any deformation including affine transformations decreases their performance. On the other hand, our methods provide invariance to both affine and isometric deformations in image space, and their performance is only affected by the quality of the input depth map. This negative effect is observed in the Kinect 1 dataset, where the depth measurements have a higher level of noise when compared to Kinect 2. It is worth mentioning that while TFEAT and Log-Polar were trained on real large-scale datasets with ground-truth correspondences from structure-from-motion settings, GeoPatch was entirely trained on simulation data. Nevertheless, it was capable of performing well on real benchmarks beyond the classic metrics, as we shall demonstrate in Section~\ref{sec:applications} with two different applications, namely, image retrieval and non-rigid tracking.

\begin{figure}[t!]
	\centering
	\begin{tabular}{c}
		\includegraphics[width=0.86\linewidth]{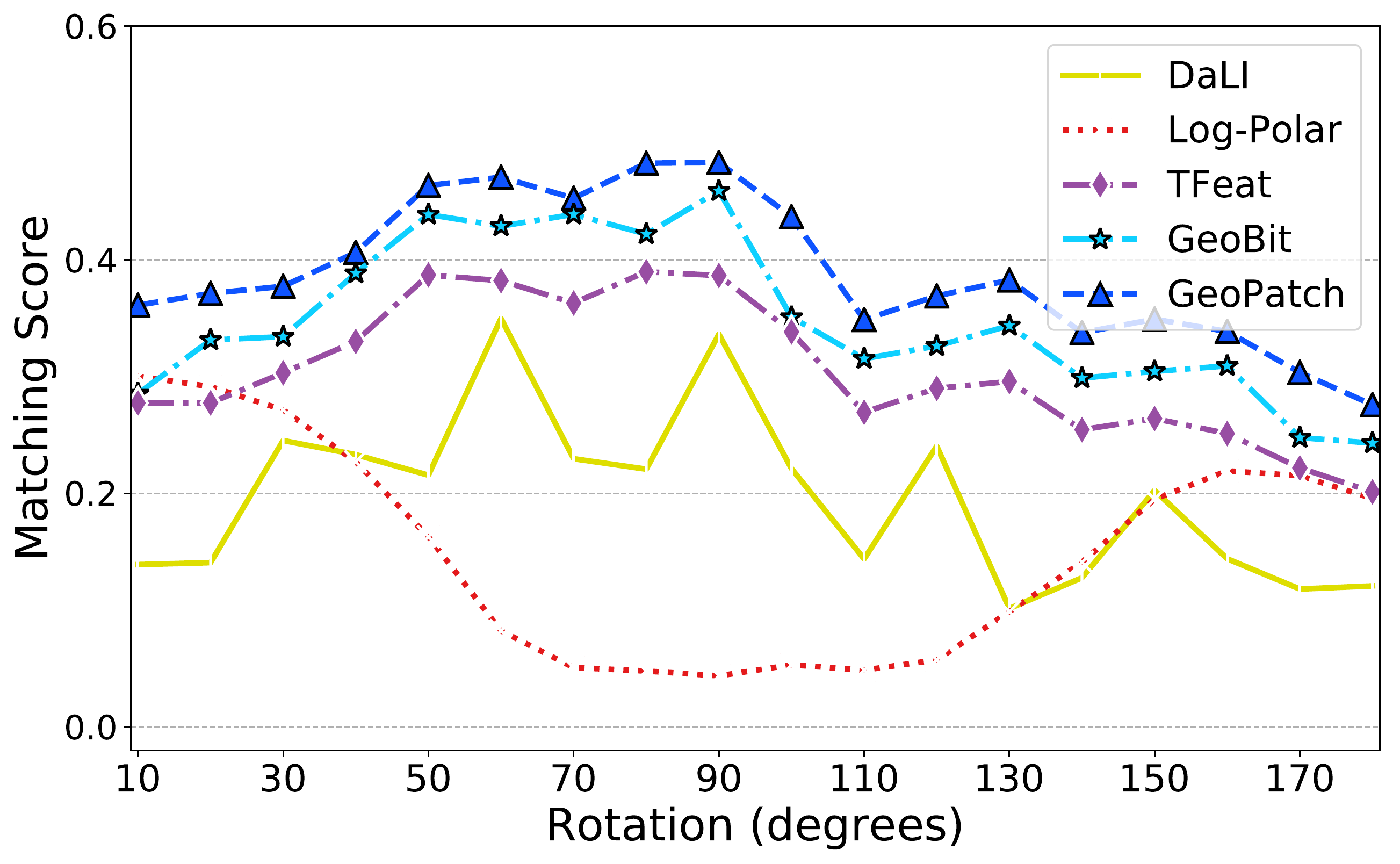} \\
		(a) Rotation invariance \\
		\includegraphics[width=0.86\linewidth]{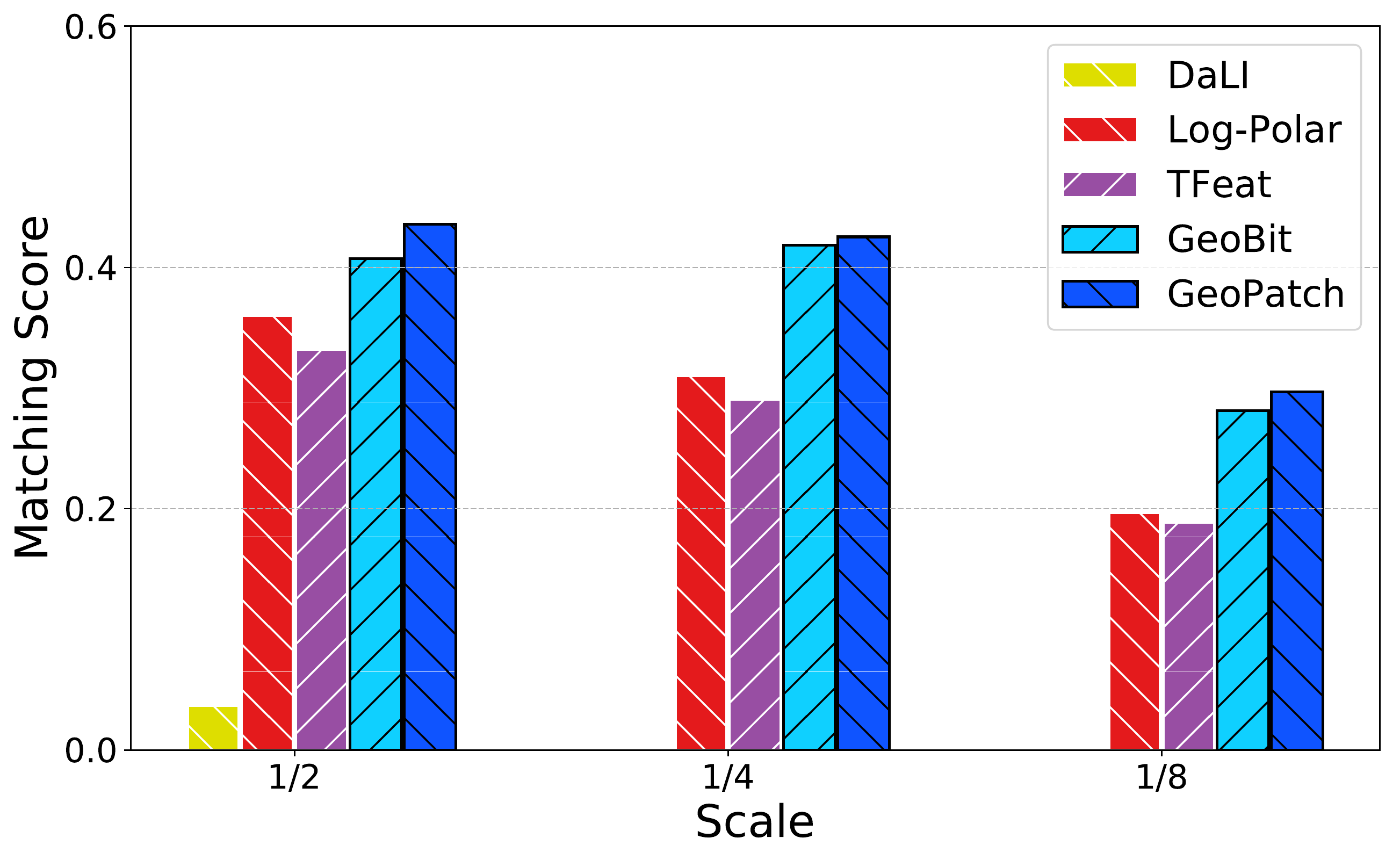} \\				
		(b) Scale invariance 
		
	\end{tabular}
	\caption{\textbf{Rotation and scale invariance.} Matching score obtained by matching SIFT keypoints considering (a) rotation and (b) scale for each target frame relative to the reference.}
	\label{fig:rotation-exp}
	
\end{figure}

\subsection{Rotation and Scale Invariance} 

We also evaluate our descriptors in terms of robustness to rotation and scale transformations. For these tests, we used the Simulation dataset, where the camera suffers in-plane rotations ranging from $0^{\circ}$ to $180^{\circ}$ degrees, using a step size of $10^{\circ}$ degrees for rotation. For the scale invariance tests,  the camera is moved backward in the Z direction to produce downscaling in image space. The rotation and scale attributes from SIFT keypoints are shared among all descriptors, with exception to DaLI and GeoBit that compare rotated versions of the descriptors, and also GeoPatch, which is invariant to rotation due to the final max pooling in the geodesic patch's angle axis.


\figurename~\ref{fig:rotation-exp} shows the matching score curves for rotation and scale transforms. The results are given by the score as a function of the rotation angle and scale. Our descriptors outperforms all methods in all frames in both rotation and scale evaluations. We can observe that our learning-based method responded with the lowest variance with respect to rotation, thanks to the pooling scheme presented, which endows the network to be rotation invariant and discriminant at the same time, without a previous orientation estimation step. This response is important since the deformations can introduce additional noise in the orientation estimation step. Under isometric surface deformations, a geodesic curve is an intrinsic property that is preserved between views. Notably, the scale invariance of our descriptors are obtained from the isometric invariance property of the geodesics contained in the depth information. We observed a decrease in Log-Polar's performance for the rotation sequences coming from an adverse effect of padding under large image in-plane rotations since all keypoints share the same orientation attribute from SIFT keypoints. Log-Polar uses a large support region due to its patch sampling scheme. 

\new{
We can observe a better performance of matching scores between $50^{\circ}$ and $90^{\circ}$ degrees for most methods in this experiment. The images used in this experiment were affected by increasing rotations and arbitrary non-rigid deformations. From visual inspection, we noticed that the frames in the range of $50^{\circ}$ - $90^{\circ}$ degrees of in-plane rotation were affected by lighter deformations than during the rotation ranges of  $0^{\circ}$ - $50^{\circ}$ and $110^{\circ}$ - $180^{\circ}$, which contained images with extreme deformations. As result, this led to a decreasing performance for most methods for frames located in the beginning and ending of the simulation sequences.
}


\subsection{Ablation Study and Processing Time}

To verify each component's contribution to our descriptors' performance, we conducted an ablation study followed by an evaluation of the parameter settings for the support region. Table~\ref{table:ablation_and_parameters} shows that the matching score increases when depth data becomes less noisy and the interpolation step slightly increases the quality of the matches. 
Moreover, despite the geodesic walk method being able to run several times (up to $60\times$) in our experiments faster than Heatflow,  both mapping functions provide the same performance in matching quality.  

Table~\ref{table:ablation_and_parameters} also shows the ablation under different conditions of depth including missing data. We can observe that using the raw depth without any preprocessing step provides the worst performance in terms of matching scores due to the high-frequency noise present in the data. The constant depth experiment was performed by using all depth as a constant value, considering the object's approximate depth relative to the camera. In this scenario, our descriptors behave like regular RGB descriptors. The pyramid smoothing achieves the best results, demonstrating that even in noisy conditions, our method can extract useful information from depth to rectify the patches. Finally, to test our methods in a scenario where there are large amounts of holes in the object’s surface, we artificially introduced holes in the depth images. The holes were generated considering the observations from the work of~\cite{cit:holedepth}, which suggests that the uncertainty of depth values from RGB-D sensors is usually higher in the edges of objects, which are also often edges in the intensity image. 
We  tested a simple filling algorithm that iterates over the image and replaces invalid depth with the last seen valid depth values, and a more elaborate method that detects the holes and uses inverse distance weighting interpolation scheme to fill the holes according to the valid pixels around the holes. The results indicate that the interpolation scheme provides modest performance gains in the matching scores.

Finally, we evaluate the impact of the strategies adopted to achieve rotation invariance for GeoPatch in the presence (\textit{chambre\_rot} dataset) and absence (\textit{Shirt3} dataset) of camera rotation. The use of max-pooling in the angle-axis of the output tensor before the fully-connected layer in GeoPatch increases the matching quality from $0.150$ (no invariance) to $0.335$ of matching score for rotation transformations without strongly diminishing the matching score in the absence of camera rotations.

\begin{table}[t!]
	\centering
	\caption{\textbf{Ablation study.} We report the average matching scores in \textit{Shirt3} and \textit{toucan\_medium} datasets.}
	\resizebox{0.7\columnwidth}{!}{%
	\begin{tabular}{lcc}
		
		\toprule      
		\textbf{Experiment} &  \textbf{GeoBit} & \textbf{GeoPatch} \\ 
		\toprule
		Raw Depth   & $0.263$ &$0.261$ \\		
		Constant Depth & $0.321$ & $0.313$  \\
		Depth Smoothing   & $0.328$ &$0.342$ \\	
		\midrule
		Heatflow  & $0.328$ & $-$  \\
		Geodesic Walk   & $0.328$ &$-$ \\        
		\midrule
		
		Simple Fill  & $0.324$ & $0.337$  \\
		Interpolation   & $0.327$ &$0.341$ \\ 

		\bottomrule
		
	\end{tabular}
	
	}      
	\label{table:ablation_and_parameters}
\end{table}

\begin{table}[t!]
	\centering
	\caption{\textbf{Processing time and descriptor size.} Time in seconds of each step for the descriptors considering $\mathbf{250}$ keypoints, and size in bytes of each descriptor.}
	\resizebox{1\columnwidth}{!}{%
		\begin{tabular}{lccrrr}
			
			\toprule      
			\textbf{Method} &  \textbf{Size} & \textbf{$f(.)$ map} & \textbf{Extraction} & \textbf{Matching} & \textbf{Total}\\ 
			\toprule
			Log-Polar  & $512$ & $-$    & $0.072$     & $0.949$ & $1.021$  \\
			TFeat   & $512$ &$-$    & $0.012$     & $1.002$ & $1.014$  \\        
			\midrule
			DaLI  & $105,600$ & $-$    & $112.95$     & $61.62$ & $174.57$  \\
			GeoBit  &$1,024$ & $0.534$    & $0.074$     & $0.531$ & $1.139$  \\    
			GeoPatch  & $512$ &$0.534$    & $0.009$     & $1.033$ & $1.576$  \\      
			\bottomrule
			
		\end{tabular}
		
	}      
	\label{table:speed}

\end{table}

\begin{figure*}[t!]
	\centering
	\begin{tabular}{cc}
		\includegraphics[width=0.63\linewidth]{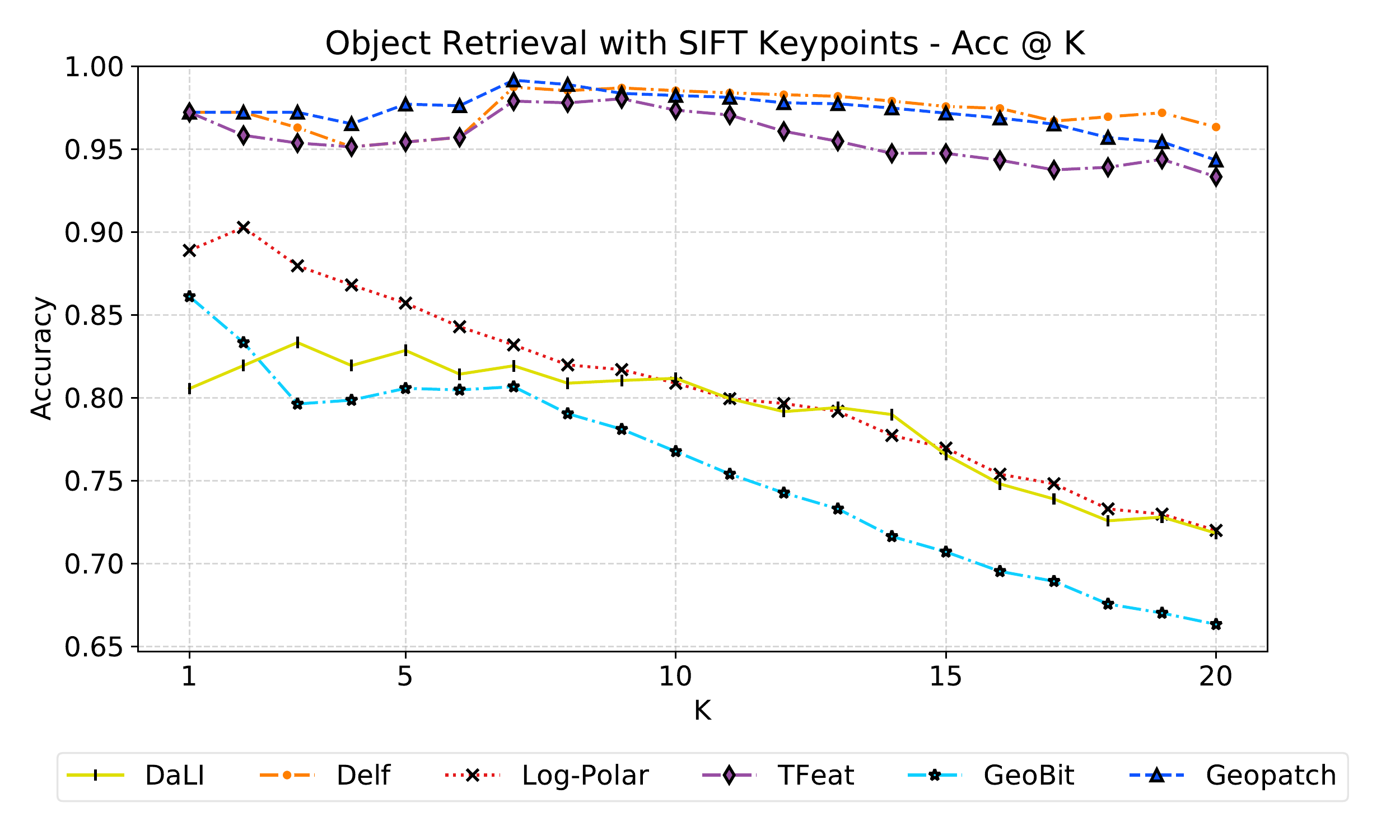} &
		\includegraphics[width=0.26\linewidth]{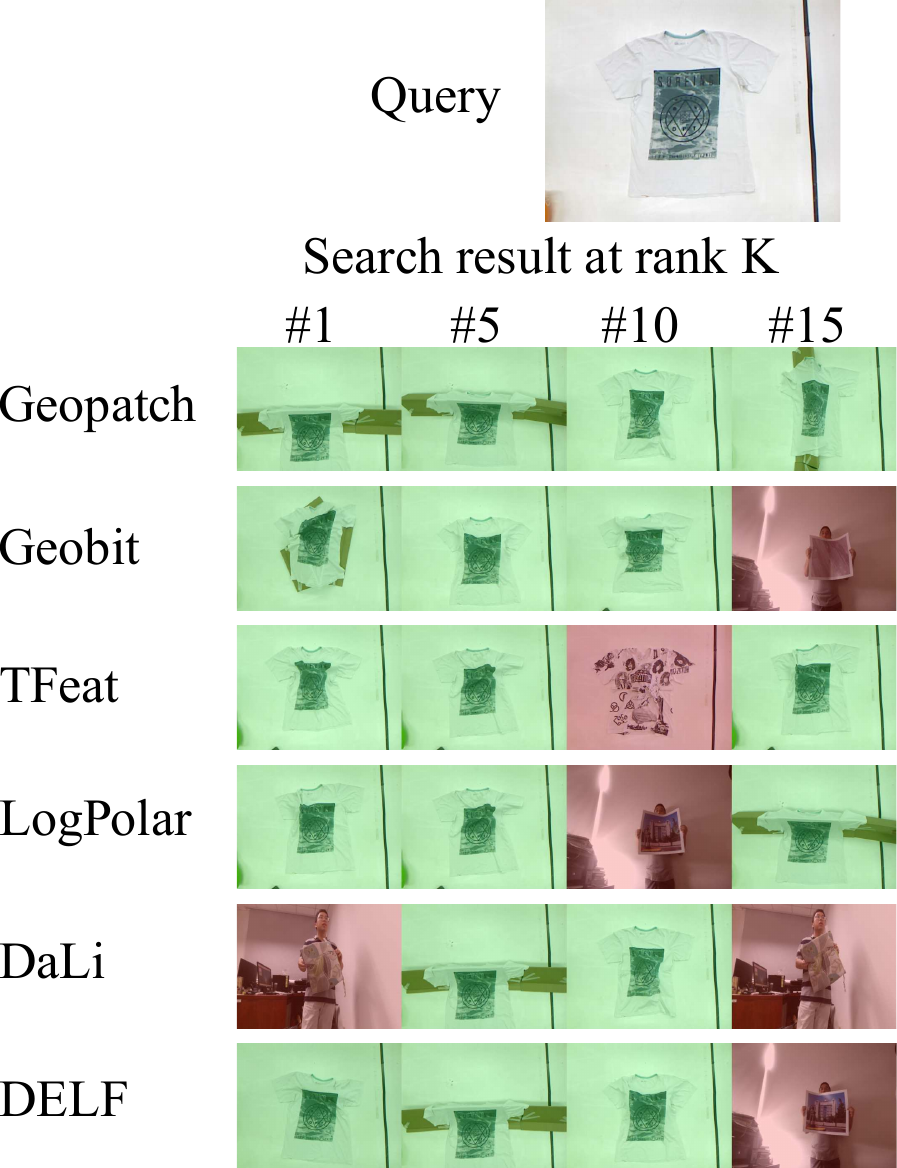} \\				
		
	\end{tabular}
	\caption{\new{{\bf Object retrieval results.} {\it Left}: Average accuracy relative to top $K$ results with different descriptors using SIFT keypoints and considering images from all datasets. {\it Right}: An example of the image retrieval for each descriptor.}}
	\label{fig:retrieval_acc}
	
\end{figure*}

Table~\ref{table:speed} shows the required time of each step for the descriptors. The code was executed on a set of $250$ keypoints, images with the resolution of $640 \times 480$ running on an Intel (R) Core (TM) i7-7700 CPU @ 3.60 GHz and a GTX 1060 GPU. The matching time refers to the brute-force matching done in CPU. 
Both of our descriptors were, on average more than $100$ times faster than DaLI, which shows state-of-the-art performance in matching regarding the description of deformable objects.
Regarding our previous strategy to extract geodesic distances, our current method runs several times faster than Heatflow strategy. While using Heatflow takes $33.263$ seconds, our new strategy runs at almost $2$ frames per second, achieving a speedup of approximately $60$ times for computing the geodesics. The learning approaches run the descriptor extraction on GPU, which results in considerable speed-up in extraction time. GeoBit may be a better choice when no GPU is available since it requires lower computational effort and runs entirely on CPU, and allows fast descriptor matching using the Hamming distance.

\section{Applications} \label{sec:applications}

Aside from the matching task, we also evaluated our descriptors on two real-world applications: object retrieval and tracking deformable objects. We detected the $2,048$ most salient SIFT keypoints per image as described in Section~\ref{sec:results}. 
In the applications, we also consider as baselines two recent state-of-the-art methods, namely DELF~\citep{cit:DELF} and R2D2~\citep{revaud2019neurips}. Please note that their provided implementation did not allow the evaluation on SIFT keypoints as for the other methods. This limitation prevents including DELF and R2D2 matching scores on our dataset, where all descriptors are evaluated with the same set of keypoints.


\subsection{Object Retrieval}

To compare the descriptors in a retrieval task, we implemented an object retrieval method based on the Bag-of-Visual-Words approach. 
The retrieval application is important to test different aspects of the descriptors, such as the distribution of the descriptor space induced by the descriptor method itself. 
In order to evaluate these properties, we built a small dictionary of $10$ representative centroids by employing the K-medoids clustering for all descriptors. We chose K-medoids to be able to seamlessly use the same method for both binary and floating-point descriptors, and also for a fair comparison among them. The visual dictionary was built on a sample of $10{,}000$ randomly chosen points considering all the datasets. Since the original images contain more than the object itself, we used a mask to detect SIFT keypoints only in the object region for all images in the database. 
After building a dictionary for each descriptor method, an image is globally described by building a frequency vector, where each bin corresponds to a visual word. For each feature, its visual word representation is computed by finding the nearest centroid in the dictionary, and its count is incremented in the respective bin in the frequency vector.

The database used for the object retrieval task is the result of the union of four datasets: {\it Simulated}, {\it Kinect 1}, {\it Kinect 2}, and {\it DeSurT}. After constructing the global descriptor vector for all images, the retrieval experiment consists in choosing the undeformed image as the query, while all images from all datasets compose a single large database of global descriptors. 
As a metric of comparison, we use the average accuracy of the retrieval over the top $K$ nearest neighbors (Acc @ $K$), using a k-NN search in the frequency vector space. The accuracy is determined by the number of correct object classes it retrieves over the top search results.
\figurename~\ref{fig:retrieval_acc} shows the result for all descriptors. One can see that GeoPatch and DELF achieved the best performance among all competitors. GeoBit, on the other hand, performed worse than GeoPatch mostly due to the increased ambiguity in the large retrieval database from the $16$ computed descriptors per keypoint on different orientations.
These results also show that, even GeoPatch being designed and trained to extract distinctive features (not discriminative features),
it was capable of presenting competitive results compared to DELF, which is tailored specifically for image retrieval. For DELF, we employ DELF detection and description steps from ~\cite{cit:DELF}, letting it detect its own keypoints optimized for image retrieval. These results concur with~\cite{cit:varma2007} and~\cite{cit:chatfield2011devil} papers, where the authors observed a clear trade-off between extracting invariant features versus discriminative features. While an invariant descriptor can provide better matches, their discriminative power, which is better for classification tasks, is diminished. 

%% file: tracking.tex
\subsection{Deformable Surface Tracking}

In this section, we present the performance of our descriptors when applied to track a region-of-interest of different textured meshes, subjected to large rotations, scale changes, variations on illumination, and strong non-rigid deformations. In this task, given a template patch in the reference image, the goal is to estimate the warp to follow the template on subsequent frames. 

From the previous image matching experiments, we selected the three best competitors in addition to our descriptors: DaLI, Log-Polar, and TFeat. Following the image retrieval task, SIFT detector was used to select $2{,}048$ keypoints in the images.  In addition, we also evaluate the state-of-the-art joint detection and description methods DELF~\citep{cit:DELF} and R2D2~\citep{revaud2019neurips}, using their own detected $2{,}048$ keypoints, for tracking deformable surfaces. For each descriptor, we computed the matching distance matrix of all visible keypoints, within the template region-of-interest, and performed the correspondences between keypoints using the smallest distances. The template tracking, \ie, the registration between the template (source image) and each current frame (target image), was performed using a Deformable-Affine Thin-Plate Spline (TPS) warp model. To reduce the effects of outliers in the performance of all descriptors, we adopted a RANSAC outlier rejection strategy~\citep{ransacTPS-eccv12} to filter out outliers. The coordinates of keypoints sets were normalized accordingly to the preprocessing discussed by~\cite{ransacTPS-eccv12}.

In order to measure the tracking accuracy quantitatively, we computed the state-of-the-art metric \textit{Learned Perceptual Patch Similarity} distance (LPIPS)~\citep{zhang2018perceptual} to estimate the visual similarity between the tracked patches and the template. For a more detailed performance assessment of the matches for each descriptor, we also present the average ratio of inlier matches and the matching scores over the selected correct matches found by RANSAC using the ground-truth correspondences from SIFT keypoints. Therefore, we present both the average accuracy from RANSAC (inlier rate) in the tracking, matching scores from the tracking (which indicates how many correspondences selected by RANSAC are correct) and the LPIPS similarity distance. These quantitative results are shown in Table~\ref{table:tracking_average} for the three most challenging sequences \textit{DeSurT}, \textit{Kinect 1}, and \textit{Simulation}. As it can be noticed, GeoPatch and GeoBit presented the highest inlier rates and matching score. 
Likewise, both geodesic-aware descriptors presented the highest perceptual similarities between the template and tracked regions overall. 
The results show that our descriptors achieved the best performance even when compared with descriptors using their own keypoints such as R2D2 and DELF in the tracking evaluation. GeoPatch presented higher inlier rates and matching scores than all descriptors, while both GeoBit and GeoPatch displayed the highest similarities overall.

\begin{table}[t!]
	\setlength{\tabcolsep}{2pt}
	
	\centering
	\caption{{\bf Evaluation of the tracking application}. Average values on all datasets. Best in bold, second-best in italic, and * indicates descriptors computed on their own detected keypoints.}
	\label{table:tracking_average}
	
	\resizebox{0.9\linewidth}{!}{%
		
		\begin{tabular}{@{}lccccc@{}}
			\toprule 
			& {{\bf Inliers RANSAC}}  &&  { {\bf Matching Score} } && { {\bf LPIPS} }  \\ 
			\midrule

			DaLI     & $ 0.32 $ && $ 0.31 $ && $ 0.45 $\\			
			DELF*    & $ 0.25 $ && $ - $ && $ 0.40 $\\			
			Log-Polar& $ 0.36 $ && $ 0.36 $ && $ 0.33 $\\			
			R2D2*    & $ \mathit{0.48} $ && $ - $ && $ 0.36 $\\			
			TFeat    & $ 0.31 $ && $ 0.37 $ && $ \mathit{0.29} $\\	
			\midrule		
			GeoBit   & $ 0.46 $ && $ \mathit{0.42} $ && $ \mathbf{0.27} $\\			
			GeoPatch & $ \mathbf{0.49} $ && $ \mathbf{0.45} $ && $ \mathit{0.29} $\\

			\bottomrule 
		\end{tabular}	
	}
\end{table}

\begin{figure}[t!]
\centering	
\small 
\includegraphics[width=0.88\linewidth]{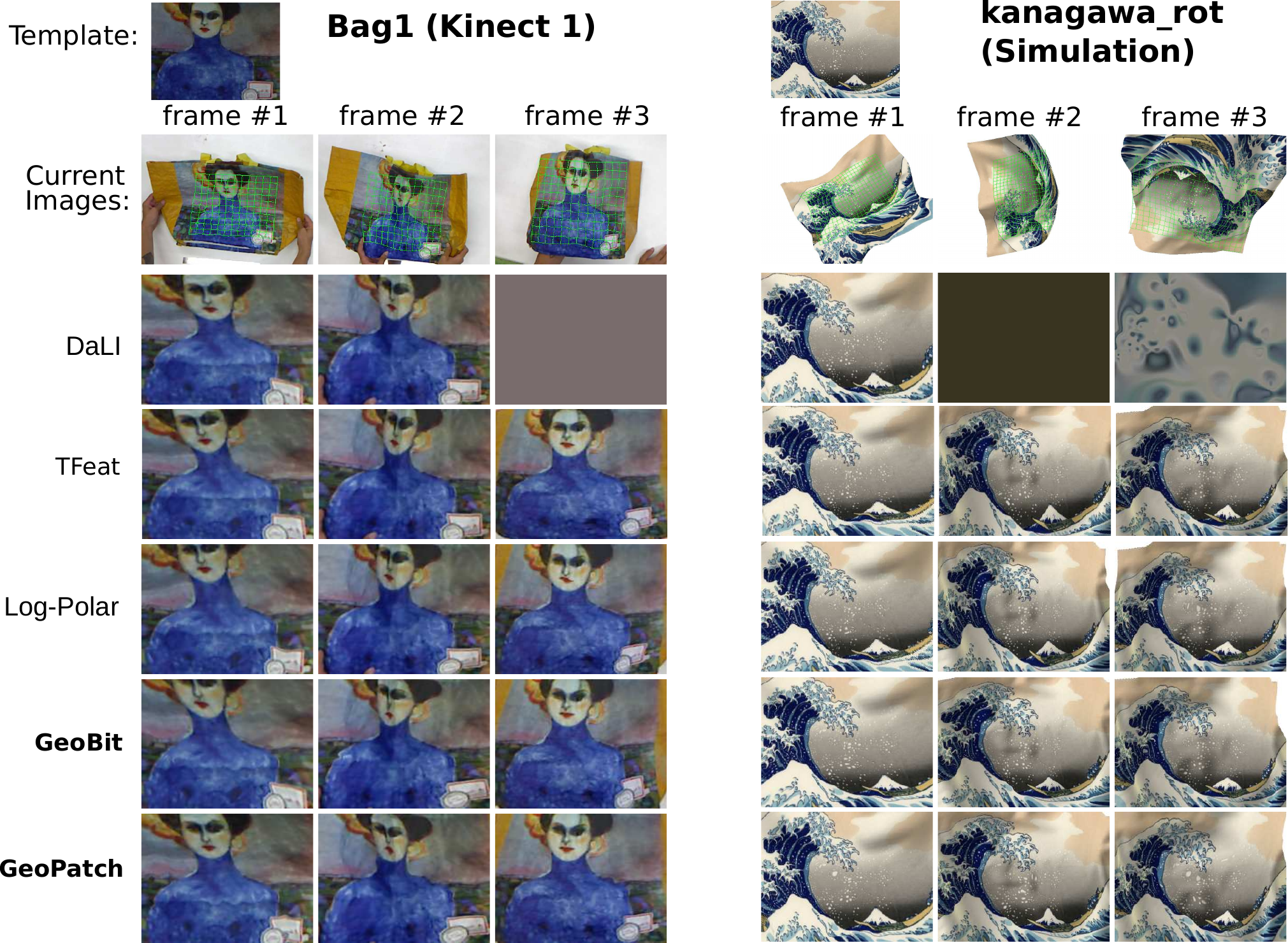}
\caption{{\bf \new{Deformable tracking visual results}}. Sequences of the test set containing strong surface deformations, orientation, and scale changes. The tracked template region is highlighted by the green grid in the first row of each sequence.}
\label{fig:tracking}
\end{figure}

Qualitative results of the tracking are shown in \figurename~\ref{fig:tracking} with sequences from our proposed non-rigid dataset and from DeSurT. Although most descriptors were capable of handling viewpoint and illumination changes, only the proposed geodesic-aware descriptors were capable of handling frames with strong surface deformations and scale changes. These effects are illustrated in \textit{chambre\_scale} and \textit{kanagawa\_rot} results. Our descriptors are also robust to  illumination changes and diffuse lighting induced by these deformations, as it can be noticed in the sequences. 
Together these results suggest the tracking of the regions-of-interest using GeoPatch and GeoBit presents better stability and consistency for both qualitative and quantitative metrics. Please check the supplementary material of this submission for an overview and additional qualitative results of the applications.

%% file: conclusion.tex
\section{Conclusions}\label{sec:concl}

In this paper, we addressed the problem of extracting local features that are isometric invariant. We have proposed a methodology for constructing scale, non-rigid deformation, and rotation invariant descriptors based on intrinsic surface properties. We designed two descriptors, a binary hand-crafted and a learning-based approach.  
Our experiments showed that our methods are robust to isometric deformation, rotation, and scale changes in image space. 
Besides the experiments showing that our descriptors outperformed all others in terms of standard metrics used to evaluate local descriptors, we also presented several experiments that demonstrate the benefits of using our methods, including real application tasks in object retrieval and tracking.

Our results take a step further towards combining photometric and geometrical information to render higher performance on local patch description. This work extends the conclusions of~\cite{tombari2011icip},~\cite{nascimento2012iros}, and~\cite{nascimento2019iccv}, where the combined use of intensity and shape information can provide invariance and distinctiveness, consequently improving the quality of keypoint matching. Thus, with the rapid progress being made in the production of multimodal sensors and the availability of inexpensive devices, it is of utmost importance to foster efficient methods that can extract and manipulate the invariance properties taking into account both sources of information.

A limiting factor of our methods when compared to RGB descriptors is the depth requirement, which is not always available. Regardless of this limitation, rapid progress is being made in estimating depth from monocular videos, and our methods to create descriptors may be benefited by these advances to run in RGB images. 
Finally, when extreme deformation arises on a surface, there is still a physical limitation in both the depth and image sensor of a camera from sampling measurements from the surface, which results in interpolation artifacts.

\section*{Acknowledgments}
The authors would like to thank CAPES (\#88881.120236/2016-01), CNPq, and FAPEMIG for funding different parts of this work. R. Martins was also supported by the French National Research Agency through grants ANR MOBIDEEP (ANR-17-CE33-0011), ANR CLARA (ANR-18-CE33-0004) and by the French Conseil Régional de Bourgogne-Franche-Comté.

%% file: appendix.tex
\newpage

\appendix 

\section{Supplementary Material}

We provide in this appendix the supplementary material to our paper \textit{Learning Geodesic-Aware Local Features from RGB-D Images}, submitted to \textit{Computer Vision and Image Understanding}. We present more detailed qualitative results of the metrics shown in \figurename~8, 9 and Table IV of the manuscript. We also further describe the dataset of non-rigid objects, and the developed intensity-based registration approach to build correspondences between the deformable images.

\begin{figure*}[b]
	\centering
	\includegraphics[width=0.93\textwidth]{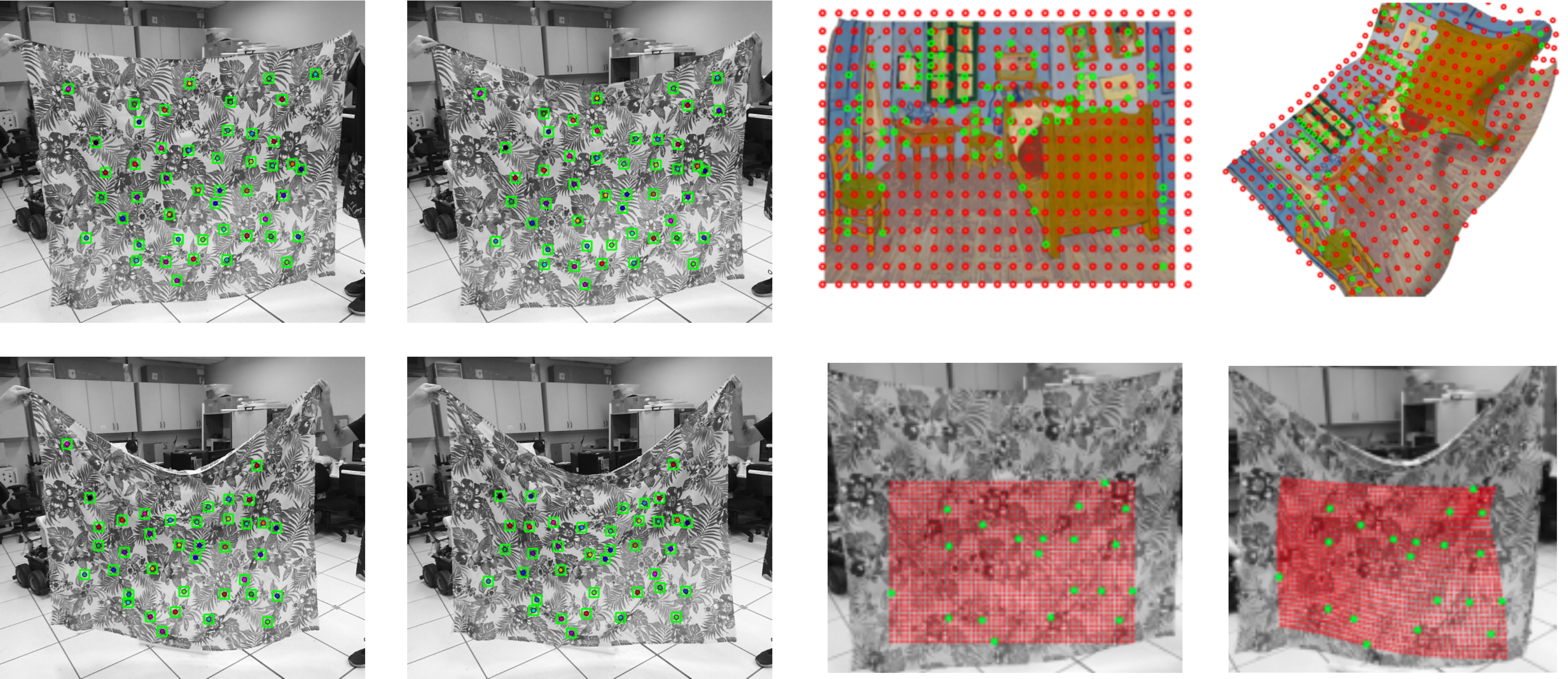}
	\caption{{\bf Ground-truth correspondences}. A mocap and manual annotation are used to estimate pixelwise-accurate landmark tracking in the images (green squares on the left plots). The right plots illustrate the estimation of the accurate ground-truth deformation model. Starting from the sparse set of accurate ground-truth annotated correspondences (indicated by the green points) as initialization to the deformation, we refine the deformation by registering the source and template images using a dense set of control points regularly sampled in the reference frame (shown in red). The deformation model is stored and used later to compute the matching scores for the independently detected keypoints for each image pair.}\label{fig:matches_mocap}   
\end{figure*}

\subsection{Dataset Dense Correspondence Generation}

Generating accurate data and assessing the quality of correspondence in the presence of non-rigid deformations is quite difficult, even for controlled experiments. Since there is no direct global geometric transformation between frames, such as the homography in the case of planar surfaces, it is hard to accurately establish pixelwise-accurate ground-truth correspondences in a large amount of independent detected keypoints across the images. In this context, we started by building a coarse set of accurate correspondences from MoCap and manual annotations (as shown by the green points in \figurename~\ref{fig:matches_mocap}). For assessing the performance of the descriptors closer to real-world conditions (large number of keypoints and detected independently), we then have developed a robust intensity-based registration approach to estimate a ground-truth deformation model for each image pair. The goal of the registration is to establish a thin-plate-spline (TPS) deformation model of densified keypoints, as shown in \figurename~\ref{fig:matches_mocap} (the initial coarse annotated keypoints are displayed in green and the sampled denser points in a regular grid in red). Since the TPS model, in general, tends to have larger errors in regions far from the control points, it is not possible to simply apply a TPS with a sparse set of correspondences and use them as ground-truth directly. In this sense, a first deformation model using TPS is estimated through the sparse set of annotated correspondences, which will be close to global optima with respect to some photometric error cost (we used structural similarity (SSIM) metric in the optimization also considering a multi-resolution Gaussian pyramid of four levels). The control points are densified by regularly sampling points in the source image and the deformation model parameters are then photometrically optimized via gradient descent using the PyTorch framework. In this way, we obtain a refined deformation model between each image pair that can be used to establish correspondences of detected keypoints reliably.

Using this strategy, we build a benchmark composed of two datasets of real-world objects acquired with two different RGB-D sensors, and an additional simulated dataset with thousands of RGB-D images of deforming objects, all with ground-truth correspondences. We provide additionally in our dataset, the estimated deformation model obtained from the registration with dense keypoints to each image pair. This model allows, in a subsequent stage, to compute the matching scores metric using a large set of independently detected keypoints per image, such as from SIFT used in the provided experiments. All datasets, descriptors and the simulation framework will be made publicly available to the community at \url{https://www.verlab.dcc.ufmg.br/descriptors}.

\balance

\subsection{Rotation Ablation}

\begin{table}[t!]
	\centering
	\caption[Rotation invariance analysis.]{\textbf{Rotation invariance analysis.} We evaluate the impact of the strategies adopted to achieve rotation invariance for GeoPatch in the presence (\textit{chambre\_rot} dataset) and absence (\textit{Shirt3} dataset) of camera rotation. Results are reported in average matching scores metric.}
	\begin{tabular}{lcc}
		
		\toprule      
		\textbf{Experiment} &  \textbf{\textit{Shirt3}} & \textbf{\textit{chambre\_rot}} \\ 
		\toprule
		
		No Rot.  & $0.355$ & $0.150$  \\
		MaxPool Angle-axis   & $0.335$ &$0.342$ \\ 
		Data Augment.   & $0.324$ &$0.335$ \\
		
		\bottomrule
		
	\end{tabular}
	
	\label{table:ablation_rot}
\end{table}

Table~\ref{table:ablation_rot} shows that the use of max-pooling in the angle-axis of the output tensor before the fully-connected layer in GeoPatch improves the matching quality for rotation transformations (\textit{chambre\_rot} sequence) without strongly diminishing the matching score in the absence of camera rotations (\textit{Shirt3} sequence).

Regarding the rotation invariance, we performed a detailed assessment of three strategies. In the first one, we did not consider rotation invariance and used it as a reference implementation. In the second strategy, we performed data augmentation by shifting the patches in the horizontal axis in a circular fashion by random offsets (from 0 to the patch width) during training in an attempt of achieving rotation invariance. The third strategy consisted of pooling the output tensor in the horizontal axis (angle axis in the geodesic patch) before the fully-connected layer. According to the results shown in Table~\ref{table:ablation_rot}, we can observe that the pooling strategy works slightly better; thus, we adopted it in our implementation.

\subsection{Additional Qualitative Results}

\subsubsection{Content-Based Image Retrieval}

We also provide some additional qualitative visual results from the object retrieval and tracking applications. \figurename~\ref{fig:retrieval_acc} shows some qualitative retrieval examples for the top six descriptors for two query images. One can see that GeoPatch achieved the best performance among all competitors along with DELF. GeoBit, on the other hand, performs worse than GeoPatch. This unexpected result can be explained by the fact that GeoBit computes $16$ descriptors on different orientations and uses the smallest distance between the rotated versions of two keypoints to incorporate rotation invariance. This strategy works well when comparing pairs of images since there is a smaller probability that a rotated patch that does not correspond to the same physical keypoint will minimize the distance among all possible rotations. However, when a large number of comparisons are considered, such as in the retrieval application, such an event can happen more frequently, resulting in more ambiguity, ultimately decreasing the accuracy. These results also show that, even GeoPatch being designed and trained to extract distinctive features (not discriminative features), it was capable of presenting competitive results compared to DELF, which in addition used its own detected keypoints for performing the retrieval.

\subsubsection{Non-Rigid Tracking} 
\begin{figure*}[tb!]
	\centering
	\includegraphics[width=0.83\linewidth]{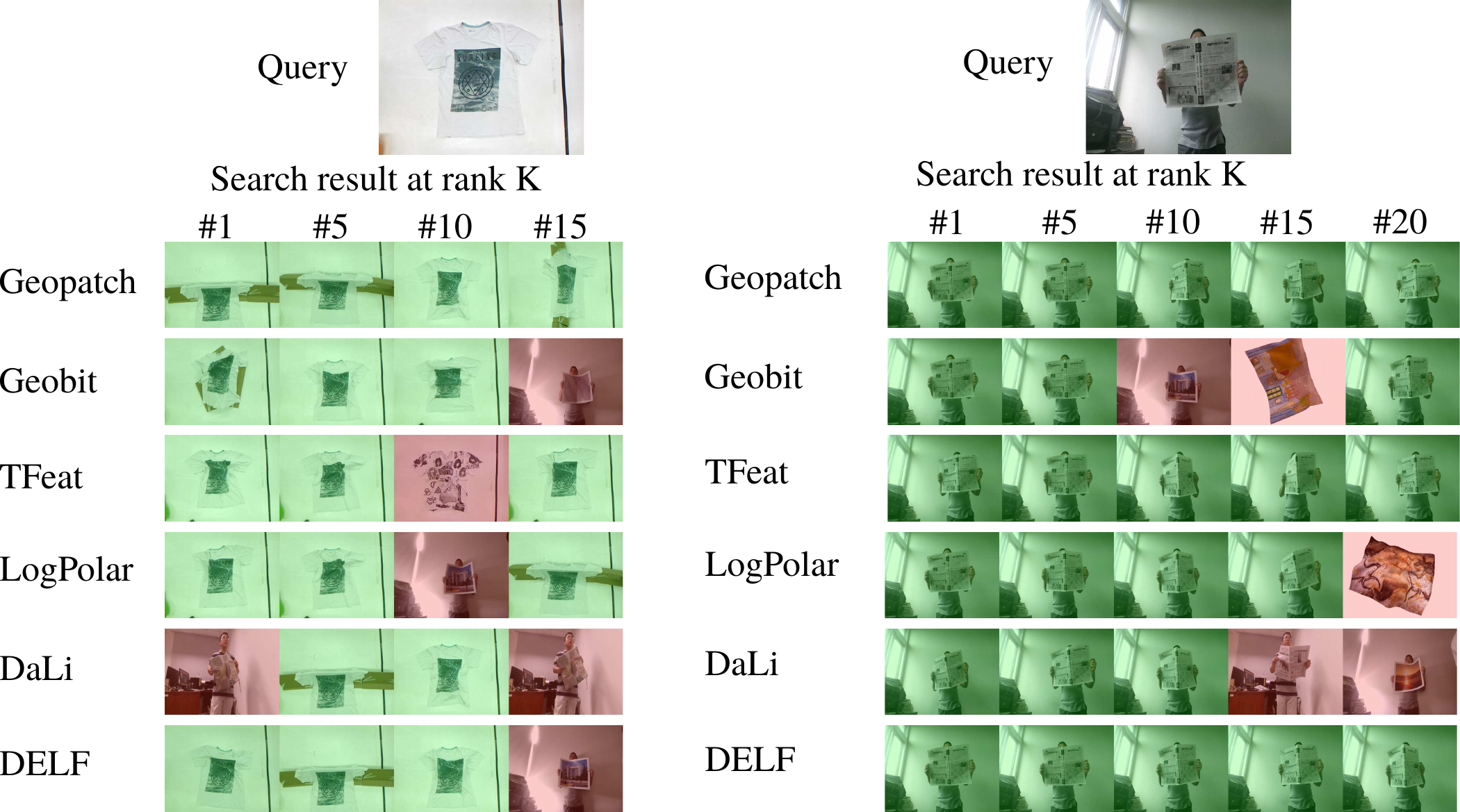}
	\caption{{\bf Qualitative results of the object retrieval application.} Two examples of the image retrieval for each descriptor.\vspace*{0.7cm}}	\label{fig:retrieval_acc}	
	
	\centering
	\includegraphics[width=0.87\linewidth]{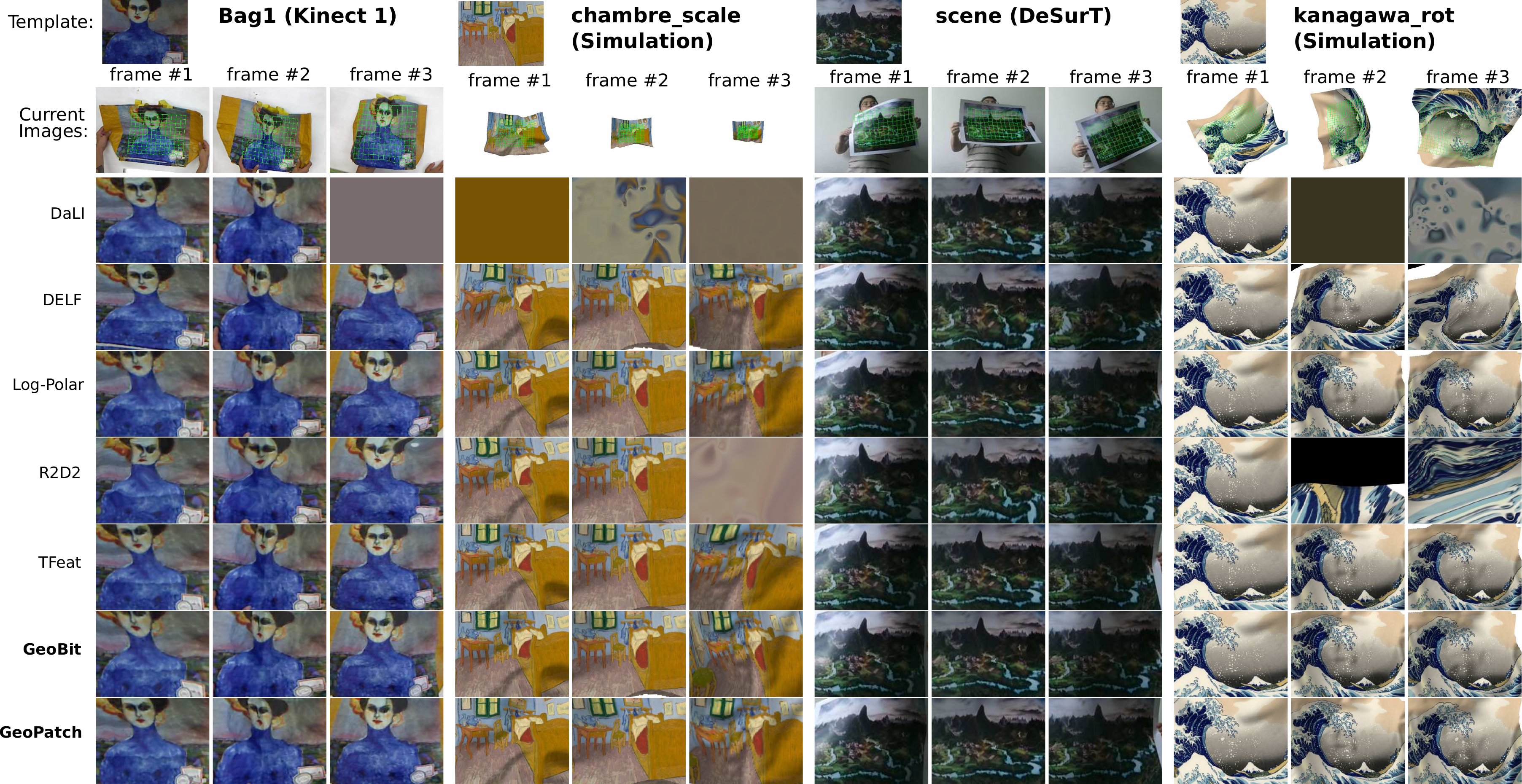}
	\caption{{\bf Deformable tracking visual results}. These sequences illustrate the test set used with detected SIFT keypoints containing mild to strong surface deformations, in addition to illumination, orientation, and scale changes. The tracked template region is highlighted by the green grid in the first row of each sequence. Notice the proposed geodesic-aware descriptors performed well for the scenes with strong deformations and scale changes.}
	\label{fig:tracking}
	
\end{figure*}

Some additional visual qualitative results of the tracking are shown in \figurename~\ref{fig:tracking} with sequences from our proposed nonrigid dataset and from DeSurT. We track the template image in the sequences over time from the computed descriptors as presented in the quantitative metrics of the paper in Table IV and \figurename~8. Although most descriptors were capable of handling viewpoint and illumination changes (as illustrated in \figurename~\ref{fig:tracking} for the sequence \textit{scene} from DeSurT), only the proposed geodesic-aware descriptors were capable of handling frames with strong surface deformations and scale changes. These effects are illustrated for the results in \textit{chambre\_scale} or \textit{kanagawa\_rot} sequences. The geodesic-aware descriptors are also robust to strong illumination changes induced by these deformations (surface not respecting the Lambertian hypothesis) and by small specular reflections, as it can be noticed in sequences \textit{scene} and \textit{chambre\_scale}.

\begin{figure}[b!]
	\centering
	\includegraphics[width=0.45\columnwidth]{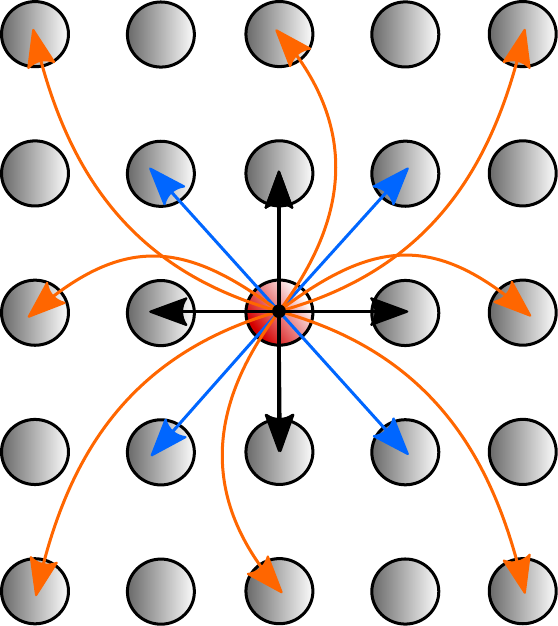}
	\caption{{\bf Connectivity of the particles.} Each edge represent a distance constraint that has to be satisfied. In the simulation, the constraint satisfaction step ensures that the Euclidean distance of each particle connected by a constraint remains constraint, to enforce isometric deformations.}
	\label{fig:particles}	
\end{figure}

\subsection{Simulation Framework}


Since it is difficult to obtain accurate ground-truth correspondences and depth of non-rigid surfaces,  we have implemented a simulation framework in OpenGL that renders a grid of particles that deforms over time. This data can be used to train deep learning models or evaluate the methods under diverse transformations. Rendering is performed using native OpenGL, which is efficient and scalable; thus, we can inexpensively generate thousands of simulated sequences.

In our simulated environment, the objects are composed by a grid of $M \times N$ particles having mass in 3D space and initially lie on a planar surface in 3D. In our implementation, $M = 80$ and $N = 60$. Each particle has a position $\mathbf{p} \in \mathbb{R}^3$ and is connected to its immediate neighbors, as demonstrated in Figure~\ref{fig:particles}. Newton's second law is applied to accumulate the acceleration vector $\mathbf{a}$ of the particle at a given instant when a force vector $\mathbf{f}$, like wind and gravity, is applied:
\begin{equation}
\mathbf{a} = \mathbf{a} + \mathbf{f}/m,
\end{equation}
\noindent where the scalar $m$ is the mass of the particle.

Then, Verlet integration is applied to translate the acceleration into velocity. Let $\mathbf{p}_{i-1}$ be the position of the particle in the last timestep $i-1$. Given the current's particle position $\mathbf{p}_i$ we can compute the next position $\mathbf{p}_{i+1}$ with the following equation:

\begin{equation}
\mathbf{p}_{i+1} = \mathbf{p}_i + (\mathbf{p}_i - \mathbf{p}_{i-1})(1 - \delta) + \mathbf{a} \Delta t^2
\end{equation}

\noindent where $\delta$ is a damping factor that accounts for air resistance, and $\Delta t$ is the constant time unit passed between two simulation steps ($\delta = 0.01$ and $\Delta t^2 = 0.175$ in our implementation). After each step, the acceleration term $\mathbf{a}$ is reset to the null vector, since it has been integrated into velocity.
After applying the motion model, a constraint satisfaction optimization is performed for each particle. The constraint satisfaction step seeks to enforce constant distance of neighboring particles according to their connectivity in the grid. Thus, for each constraint connecting particle $j$ to particle $k$, given the target distance constraint $\eta_{j,k}$, we compute a correction vector $ \Delta \mathbf{e}_{j,k}$ as follows: 
\begin{equation}
\Delta \mathbf{e}_{j,k} = (\mathbf{p}_j - \mathbf{p}_k)\dfrac{\eta_{j,k}}{\|\mathbf{p}_j - \mathbf{p}_k\|} , 
\end{equation}
\noindent and apply it to both particles $\mathbf{p}_j= \mathbf{p}_j + 0.5\Delta \mathbf{e}_{j,k} $ and $\mathbf{p}_k = \mathbf{p}_k -0.5\Delta \mathbf{e}_{j,k}$ to enforce the target distance $\eta_{i,j}$ between them. In our framework this iterative process is repeated $25$ times, which is enough to converge.

To ensure that each simulated sequence is diverse and independent from each other, we propose to use large-scale datasets of real images obtained from the internet~\citep{cit:1dsfm} and import them as a texture to be used in the mesh defined by the grid of particles. To clean the data, we first filtered out images that contain people (for privacy reasons) by using an ensemble of image detectors and then removed the few remaining images containing people by hand.

While being deformed, the texture is rendered using diffuse illumination, causing non-linear illumination changes. Landmarks are chosen by projecting each particle onto the virtual camera and retaining corner-like features according to the Harris score. We keep the best $100$ points and use them as landmarks to compute the TPS model, which provides the groundtruth correspondence to the initial rough deformation model, which is refined by the registration using the denser sets of keypoints. To enforce rich transformations of the object, we use random inputs for the simulation parameters, including wind force and direction, illumination strength, position and number of light sources, and Gaussian noise in the image pixels. This strategy guarantee realistic chaotic behavior of deformations, in addition to non-linear illumination changes and noise in the virtual camera.

%% file: main.bbl
\begin{thebibliography}{56}
\expandafter\ifx\csname natexlab\endcsname\relax\def\natexlab#1{#1}\fi
\providecommand{\url}[1]{\texttt{#1}}
\providecommand{\href}[2]{#2}
\providecommand{\path}[1]{#1}
\providecommand{\DOIprefix}{doi:}
\providecommand{\ArXivprefix}{arXiv:}
\providecommand{\URLprefix}{URL: }
\providecommand{\Pubmedprefix}{pmid:}
\providecommand{\doi}[1]{\href{http://dx.doi.org/#1}{\path{#1}}}
\providecommand{\Pubmed}[1]{\href{pmid:#1}{\path{#1}}}
\providecommand{\bibinfo}[2]{#2}
\ifx\xfnm\relax \def\xfnm[#1]{\unskip,\space#1}\fi
\bibitem[{Alahi et~al.(2012)Alahi, Ortiz and
  Vandergheynst}]{cit:alahi2012freak}
\bibinfo{author}{Alahi, A.}, \bibinfo{author}{Ortiz, R.},
  \bibinfo{author}{Vandergheynst, P.}, \bibinfo{year}{2012}.
\newblock \bibinfo{title}{Freak: Fast retina keypoint}, in:
  \bibinfo{booktitle}{Proc. CVPR}.
\bibitem[{Bartoli et~al.(2010)Bartoli, Perriollat and Chambon}]{tps}
\bibinfo{author}{Bartoli, A.}, \bibinfo{author}{Perriollat, M.},
  \bibinfo{author}{Chambon, S.}, \bibinfo{year}{2010}.
\newblock \bibinfo{title}{Generalized thin-plate spline warps}.
\newblock \bibinfo{journal}{Int. J. Comput. Vis.} \bibinfo{volume}{88},
  \bibinfo{pages}{85--110}.
\bibitem[{Bay et~al.(2008)Bay, Ess, Tuytelaars and Gool}]{surf2008cviu}
\bibinfo{author}{Bay, H.}, \bibinfo{author}{Ess, A.},
  \bibinfo{author}{Tuytelaars, T.}, \bibinfo{author}{Gool, L.V.},
  \bibinfo{year}{2008}.
\newblock \bibinfo{title}{Speeded-up robust features ({SURF})}.
\newblock \bibinfo{journal}{Computer Vision and Image Understanding}
  \bibinfo{volume}{110}, \bibinfo{pages}{346 -- 359}.
\bibitem[{Bozic et~al.(2020)Bozic, Zollhofer, Theobalt and
  Nie{\ss}ner}]{bozic2020deepdeform}
\bibinfo{author}{Bozic, A.}, \bibinfo{author}{Zollhofer, M.},
  \bibinfo{author}{Theobalt, C.}, \bibinfo{author}{Nie{\ss}ner, M.},
  \bibinfo{year}{2020}.
\newblock \bibinfo{title}{Deepdeform: Learning non-rigid rgb-d reconstruction
  with semi-supervised data}, in: \bibinfo{booktitle}{Proc. CVPR}.
\bibitem[{Brown et~al.(2010)Brown, Hua and Winder}]{cit:brown_patches_learning}
\bibinfo{author}{Brown, M.}, \bibinfo{author}{Hua, G.},
  \bibinfo{author}{Winder, S.}, \bibinfo{year}{2010}.
\newblock \bibinfo{title}{Discriminative learning of local image descriptors}.
\newblock \bibinfo{journal}{IEEE Trans. PAMI} \bibinfo{volume}{33},
  \bibinfo{pages}{43--57}.
\bibitem[{{Calonder} et~al.(2012){Calonder}, {Lepetit}, {Ozuysal}, {Trzcinski},
  {Strecha} and {Fua}}]{brief2012tpami}
\bibinfo{author}{{Calonder}, M.}, \bibinfo{author}{{Lepetit}, V.},
  \bibinfo{author}{{Ozuysal}, M.}, \bibinfo{author}{{Trzcinski}, T.},
  \bibinfo{author}{{Strecha}, C.}, \bibinfo{author}{{Fua}, P.},
  \bibinfo{year}{2012}.
\newblock \bibinfo{title}{Brief: Computing a local binary descriptor very
  fast}.
\newblock \bibinfo{journal}{IEEE Trans. PAMI} \bibinfo{volume}{34},
  \bibinfo{pages}{1281--1298}.
\bibitem[{Cavallari et~al.(2020)Cavallari, Golodetz, Lord, Valentin,
  Prisacariu, Stefano and Torr}]{cavallari2020pami}
\bibinfo{author}{Cavallari, T.}, \bibinfo{author}{Golodetz, S.},
  \bibinfo{author}{Lord, N.A.}, \bibinfo{author}{Valentin, J.},
  \bibinfo{author}{Prisacariu, V.A.}, \bibinfo{author}{Stefano, L.},
  \bibinfo{author}{Torr, P.S.}, \bibinfo{year}{2020}.
\newblock \bibinfo{title}{Real-time {RGB-D} camera pose estimation in novel
  scenes using a relocalisation cascade}.
\newblock \bibinfo{journal}{IEEE Trans. PAMI} \bibinfo{volume}{42},
  \bibinfo{pages}{2465--2477}.
\bibitem[{Chatfield et~al.(2011)Chatfield, Lempitsky, Vedaldi and
  Zisserman}]{cit:chatfield2011devil}
\bibinfo{author}{Chatfield, K.}, \bibinfo{author}{Lempitsky, V.S.},
  \bibinfo{author}{Vedaldi, A.}, \bibinfo{author}{Zisserman, A.},
  \bibinfo{year}{2011}.
\newblock \bibinfo{title}{The devil is in the details: an evaluation of recent
  feature encoding methods.}, in: \bibinfo{booktitle}{Proc. BMVC}.
\bibitem[{Crane et~al.(2013)Crane, Weischedel and Wardetzky}]{Crane:2013:TOG}
\bibinfo{author}{Crane, K.}, \bibinfo{author}{Weischedel, C.},
  \bibinfo{author}{Wardetzky, M.}, \bibinfo{year}{2013}.
\newblock \bibinfo{title}{Geodesics in heat: A new approach to computing
  distance based on heat flow}.
\newblock \bibinfo{journal}{ACM Trans. Graph. (TOG)} \bibinfo{volume}{32},
  \bibinfo{pages}{152:1--152:11}.
\bibitem[{DeTone et~al.(2018)DeTone, Malisiewicz and
  Rabinovich}]{cit:superpoint}
\bibinfo{author}{DeTone, D.}, \bibinfo{author}{Malisiewicz, T.},
  \bibinfo{author}{Rabinovich, A.}, \bibinfo{year}{2018}.
\newblock \bibinfo{title}{Superpoint: Self-supervised interest point detection
  and description}, in: \bibinfo{booktitle}{IEEE CVPR Workshops}.
\bibitem[{Dryanovski et~al.(2013)Dryanovski, Valenti and Xiao}]{cit:holedepth}
\bibinfo{author}{Dryanovski, I.}, \bibinfo{author}{Valenti, R.G.},
  \bibinfo{author}{Xiao, J.}, \bibinfo{year}{2013}.
\newblock \bibinfo{title}{Fast visual odometry and mapping from {RGB-D} data},
  in: \bibinfo{booktitle}{Proc. ICRA}.
\bibitem[{Ebel et~al.(2019)Ebel, Mishchuk, Yi, Fua and
  Trulls}]{cit:beyondcartesian}
\bibinfo{author}{Ebel, P.}, \bibinfo{author}{Mishchuk, A.},
  \bibinfo{author}{Yi, K.M.}, \bibinfo{author}{Fua, P.},
  \bibinfo{author}{Trulls, E.}, \bibinfo{year}{2019}.
\newblock \bibinfo{title}{Beyond cartesian representations for local
  descriptors}, in: \bibinfo{booktitle}{Proc. ICCV}.
\bibitem[{Guan and Smith(2017)}]{brisks17}
\bibinfo{author}{Guan, H.}, \bibinfo{author}{Smith, W.A.},
  \bibinfo{year}{2017}.
\newblock \bibinfo{title}{{BRISKS}: Binary features for spherical images on a
  geodesic grid}, in: \bibinfo{booktitle}{Proc. CVPR}.
\bibitem[{Harris and Stephens(1988)}]{harris}
\bibinfo{author}{Harris, C.}, \bibinfo{author}{Stephens, M.},
  \bibinfo{year}{1988}.
\newblock \bibinfo{title}{A combined corner and edge detector}, in:
  \bibinfo{booktitle}{Proceedings of the Alvey Vision Conference}, pp.
  \bibinfo{pages}{23.1--23.6}.
\bibitem[{Jin et~al.(2021)Jin, Mishkin, Mishchuk, Matas, Fua, Yi and
  Trulls}]{jin2021image}
\bibinfo{author}{Jin, Y.}, \bibinfo{author}{Mishkin, D.},
  \bibinfo{author}{Mishchuk, A.}, \bibinfo{author}{Matas, J.},
  \bibinfo{author}{Fua, P.}, \bibinfo{author}{Yi, K.M.},
  \bibinfo{author}{Trulls, E.}, \bibinfo{year}{2021}.
\newblock \bibinfo{title}{Image matching across wide baselines: From paper to
  practice}.
\newblock \bibinfo{journal}{Int. J. Comput. Vis.} \bibinfo{volume}{129},
  \bibinfo{pages}{517--547}.
\bibitem[{Kokkinos et~al.(2012)Kokkinos, Bronstein, Litman and
  Bronstein}]{Kokkinos2012cvpr}
\bibinfo{author}{Kokkinos, I.}, \bibinfo{author}{Bronstein, M.M.},
  \bibinfo{author}{Litman, R.}, \bibinfo{author}{Bronstein, A.M.},
  \bibinfo{year}{2012}.
\newblock \bibinfo{title}{Intrinsic shape context descriptors for deformable
  shapes}, in: \bibinfo{booktitle}{Proc. CVPR}.
\bibitem[{Laguna et~al.(2019)Laguna, Riba, Ponsa and Mikolajczyk}]{cit:key.net}
\bibinfo{author}{Laguna, A.B.}, \bibinfo{author}{Riba, E.},
  \bibinfo{author}{Ponsa, D.}, \bibinfo{author}{Mikolajczyk, K.},
  \bibinfo{year}{2019}.
\newblock \bibinfo{title}{Key. net: Keypoint detection by handcrafted and
  learned cnn filters}.
\newblock \bibinfo{journal}{arXiv} .
\bibitem[{Leutenegger et~al.(2011)Leutenegger, Chli and Siegwart}]{brisk11}
\bibinfo{author}{Leutenegger, S.}, \bibinfo{author}{Chli, M.},
  \bibinfo{author}{Siegwart, R.}, \bibinfo{year}{2011}.
\newblock \bibinfo{title}{{BRISK}: Binary robust invariant scalable keypoints},
  in: \bibinfo{booktitle}{Proc. ICCV}.
\bibitem[{Lin and Huang(2020)}]{huang2020pami}
\bibinfo{author}{Lin, D.}, \bibinfo{author}{Huang, H.}, \bibinfo{year}{2020}.
\newblock \bibinfo{title}{Zig-zag network for semantic segmentation of rgb-d
  images}.
\newblock \bibinfo{journal}{IEEE Trans. PAMI} \bibinfo{volume}{42},
  \bibinfo{pages}{2642--2655}.
\bibitem[{Lowe(2004)}]{lowe2004ijcv}
\bibinfo{author}{Lowe, D.G.}, \bibinfo{year}{2004}.
\newblock \bibinfo{title}{Distinctive image features from scale-invariant
  keypoints}.
\newblock \bibinfo{journal}{Int. J. Comput. Vis.} , \bibinfo{pages}{91--110}.
\bibitem[{Martins et~al.(2016)Martins, Fernandez-Moral and
  Rives}]{martins2016adaptive}
\bibinfo{author}{Martins, R.}, \bibinfo{author}{Fernandez-Moral, E.},
  \bibinfo{author}{Rives, P.}, \bibinfo{year}{2016}.
\newblock \bibinfo{title}{Adaptive direct {RGB-D} registration and mapping for
  large motions}, in: \bibinfo{booktitle}{Proc. ACCV}.
\bibitem[{Mikolajczyk et~al.(2005)Mikolajczyk, Tuytelaars, Schmid, Zisserman,
  Matas, Schaffalitzky, Kadir and Van~Gool}]{cit:matchscore}
\bibinfo{author}{Mikolajczyk, K.}, \bibinfo{author}{Tuytelaars, T.},
  \bibinfo{author}{Schmid, C.}, \bibinfo{author}{Zisserman, A.},
  \bibinfo{author}{Matas, J.}, \bibinfo{author}{Schaffalitzky, F.},
  \bibinfo{author}{Kadir, T.}, \bibinfo{author}{Van~Gool, L.},
  \bibinfo{year}{2005}.
\newblock \bibinfo{title}{A comparison of affine region detectors}.
\newblock \bibinfo{journal}{Int. J. Comput. Vis.} \bibinfo{volume}{65},
  \bibinfo{pages}{43--72}.
\bibitem[{Mishchuk et~al.(2017)Mishchuk, Mishkin, Radenovic and
  Matas}]{cit:hardnet}
\bibinfo{author}{Mishchuk, A.}, \bibinfo{author}{Mishkin, D.},
  \bibinfo{author}{Radenovic, F.}, \bibinfo{author}{Matas, J.},
  \bibinfo{year}{2017}.
\newblock \bibinfo{title}{Working hard to know your neighbor's margins: Local
  descriptor learning loss}, in: \bibinfo{booktitle}{Proc. NeurIPS}.
\bibitem[{Moreno-Noguer(2011)}]{dali}
\bibinfo{author}{Moreno-Noguer, F.}, \bibinfo{year}{2011}.
\newblock \bibinfo{title}{Deformation and illumination invariant feature point
  descriptor}, in: \bibinfo{booktitle}{Proc. CVPR}.
\bibitem[{Nakajima et~al.(2019)Nakajima, Kang, Saito and
  Kitani}]{nakajima2019iccv}
\bibinfo{author}{Nakajima, Y.}, \bibinfo{author}{Kang, B.},
  \bibinfo{author}{Saito, H.}, \bibinfo{author}{Kitani, K.},
  \bibinfo{year}{2019}.
\newblock \bibinfo{title}{Incremental class discovery for semantic segmentation
  with rgbd sensing}, in: \bibinfo{booktitle}{Proc. ICCV}.
\bibitem[{Nascimento et~al.(2012a)Nascimento, Oliveira, Campos, Vieira and
  Schwartz}]{nascimento2012iros}
\bibinfo{author}{Nascimento, E.R.}, \bibinfo{author}{Oliveira, G.L.},
  \bibinfo{author}{Campos, M.F.M.}, \bibinfo{author}{Vieira, A.W.},
  \bibinfo{author}{Schwartz, W.R.}, \bibinfo{year}{2012}a.
\newblock \bibinfo{title}{{BRAND: A Robust Appearance and Depth Descriptor for
  RGB-D Images}}, in: \bibinfo{booktitle}{Proc. IROS}.
\bibitem[{Nascimento et~al.(2013)Nascimento, Oliveira, Vieira and
  Campos}]{Nascimento:Neurocomputing:2013}
\bibinfo{author}{Nascimento, E.R.}, \bibinfo{author}{Oliveira, G.L.},
  \bibinfo{author}{Vieira, A.W.}, \bibinfo{author}{Campos, M.F.M.},
  \bibinfo{year}{2013}.
\newblock \bibinfo{title}{{On the development of a robust, fast and lightweight
  keypoint descriptor}}.
\newblock \bibinfo{journal}{Neurocomputing} \bibinfo{volume}{120},
  \bibinfo{pages}{141--155}.
\bibitem[{{Nascimento} et~al.(2019){Nascimento}, {Potje}, {Martins}, {Cadar},
  {Campos} and {Bajcsy}}]{nascimento2019iccv}
\bibinfo{author}{{Nascimento}, E.R.}, \bibinfo{author}{{Potje}, G.},
  \bibinfo{author}{{Martins}, R.}, \bibinfo{author}{{Cadar}, F.},
  \bibinfo{author}{{Campos}, M.F.M.}, \bibinfo{author}{{Bajcsy}, R.},
  \bibinfo{year}{2019}.
\newblock \bibinfo{title}{{GEOBIT}: A geodesic-based binary descriptor
  invariant to non-rigid deformations for {RGB-D} images}, in:
  \bibinfo{booktitle}{Proc. ICCV}.
\bibitem[{Nascimento et~al.(2012b)Nascimento, Schwartz and
  Campos}]{nascimento2012icpr}
\bibinfo{author}{Nascimento, E.R.}, \bibinfo{author}{Schwartz, W.R.},
  \bibinfo{author}{Campos, M.F.M.}, \bibinfo{year}{2012}b.
\newblock \bibinfo{title}{{EDVD} - enhanced descriptor for visual and depth
  data}, in: \bibinfo{booktitle}{Proc. ICPR}.
\bibitem[{Ojala et~al.(1996)Ojala, Pietikäinen and
  Harwood}]{ojala1996patternrecognition}
\bibinfo{author}{Ojala, T.}, \bibinfo{author}{Pietikäinen, M.},
  \bibinfo{author}{Harwood, D.}, \bibinfo{year}{1996}.
\newblock \bibinfo{title}{A comparative study of texture measures with
  classification based on featured distributions}.
\newblock \bibinfo{journal}{Pattern Recognition} \bibinfo{volume}{29},
  \bibinfo{pages}{51 -- 59}.
\bibitem[{Ono et~al.(2018)Ono, Trulls, Fua and Yi}]{cit:lfnet}
\bibinfo{author}{Ono, Y.}, \bibinfo{author}{Trulls, E.}, \bibinfo{author}{Fua,
  P.}, \bibinfo{author}{Yi, K.M.}, \bibinfo{year}{2018}.
\newblock \bibinfo{title}{Lf-net: learning local features from images}, in:
  \bibinfo{booktitle}{Proc. NeurIPS}.
\bibitem[{Revaud et~al.(2019)Revaud, De~Souza, Humenberger and
  Weinzaepfel}]{revaud2019neurips}
\bibinfo{author}{Revaud, J.}, \bibinfo{author}{De~Souza, C.},
  \bibinfo{author}{Humenberger, M.}, \bibinfo{author}{Weinzaepfel, P.},
  \bibinfo{year}{2019}.
\newblock \bibinfo{title}{{R2D2}: Reliable and repeatable detector and
  descriptor}, in: \bibinfo{booktitle}{Proc. NeurIPS}.
\bibitem[{Rublee et~al.(2011)Rublee, Rabaud, Konolige and
  Bradski}]{rublee2011iccv}
\bibinfo{author}{Rublee, E.}, \bibinfo{author}{Rabaud, V.},
  \bibinfo{author}{Konolige, K.}, \bibinfo{author}{Bradski, G.},
  \bibinfo{year}{2011}.
\newblock \bibinfo{title}{{ORB: an efficient alternative to SIFT or SURF}}, in:
  \bibinfo{booktitle}{Proc. ICCV}.
\bibitem[{Shamai and Kimmel(2017)}]{Shamai2017GeodesicDD}
\bibinfo{author}{Shamai, G.}, \bibinfo{author}{Kimmel, R.},
  \bibinfo{year}{2017}.
\newblock \bibinfo{title}{Geodesic distance descriptors}, in:
  \bibinfo{booktitle}{Proc. CVPR}.
\bibitem[{Simo-Serra et~al.(2015a)Simo-Serra, Torras and
  Moreno-Noguer}]{simo2015dali}
\bibinfo{author}{Simo-Serra, E.}, \bibinfo{author}{Torras, C.},
  \bibinfo{author}{Moreno-Noguer, F.}, \bibinfo{year}{2015}a.
\newblock \bibinfo{title}{{DaLI}: deformation and light invariant descriptor}.
\newblock \bibinfo{journal}{Int. J. Comput. Vis.} \bibinfo{volume}{115}.
\bibitem[{Simo-Serra et~al.(2015b)Simo-Serra, Trulls, Ferraz, Kokkinos, Fua and
  Moreno-Noguer}]{simo-iccv15}
\bibinfo{author}{Simo-Serra, E.}, \bibinfo{author}{Trulls, E.},
  \bibinfo{author}{Ferraz, L.}, \bibinfo{author}{Kokkinos, I.},
  \bibinfo{author}{Fua, P.}, \bibinfo{author}{Moreno-Noguer, F.},
  \bibinfo{year}{2015}b.
\newblock \bibinfo{title}{Discriminative learning of deep convolutional feature
  point descriptors}, in: \bibinfo{booktitle}{Proc. ICCV}.
\bibitem[{Surazhsky et~al.(2005)Surazhsky, Surazhsky, Kirsanov, Gortler and
  Hoppe}]{fastmarching}
\bibinfo{author}{Surazhsky, V.}, \bibinfo{author}{Surazhsky, T.},
  \bibinfo{author}{Kirsanov, D.}, \bibinfo{author}{Gortler, S.J.},
  \bibinfo{author}{Hoppe, H.}, \bibinfo{year}{2005}.
\newblock \bibinfo{title}{Fast exact and approximate geodesics on meshes}, in:
  \bibinfo{booktitle}{ACM Trans. Graph. (TOG)}, \bibinfo{organization}{Acm}.
  pp. \bibinfo{pages}{553--560}.
\bibitem[{Teichmann et~al.(2019)Teichmann, Araujo, Zhu and Sim}]{cit:DELF}
\bibinfo{author}{Teichmann, M.}, \bibinfo{author}{Araujo, A.},
  \bibinfo{author}{Zhu, M.}, \bibinfo{author}{Sim, J.}, \bibinfo{year}{2019}.
\newblock \bibinfo{title}{Detect-to-retrieve: Efficient regional aggregation
  for image search}, in: \bibinfo{booktitle}{Proc. CVPR}.
\bibitem[{Tian et~al.(2017)Tian, Fan and Wu}]{cit:l2net}
\bibinfo{author}{Tian, Y.}, \bibinfo{author}{Fan, B.}, \bibinfo{author}{Wu,
  F.}, \bibinfo{year}{2017}.
\newblock \bibinfo{title}{L2-net: Deep learning of discriminative patch
  descriptor in euclidean space}, in: \bibinfo{booktitle}{Proc. CVPR}.
\bibitem[{Tola et~al.(2010)Tola, Lepetit and Fua}]{daisy}
\bibinfo{author}{Tola, E.}, \bibinfo{author}{Lepetit, V.},
  \bibinfo{author}{Fua, P.}, \bibinfo{year}{2010}.
\newblock \bibinfo{title}{Daisy: An efficient dense descriptor applied to wide
  baseline stereo}.
\newblock \bibinfo{journal}{IEEE Trans. PAMI} \bibinfo{volume}{32},
  \bibinfo{pages}{815--830}.
\bibitem[{Tombari et~al.(2011)Tombari, Salti and Stefano}]{tombari2011icip}
\bibinfo{author}{Tombari, F.}, \bibinfo{author}{Salti, S.},
  \bibinfo{author}{Stefano, L.D.}, \bibinfo{year}{2011}.
\newblock \bibinfo{title}{{A combined texture-shape descriptor for enhanced 3D
  feature matching}}, in: \bibinfo{booktitle}{Proc. ICIP}.
\bibitem[{Tran et~al.(2012)Tran, Chin, Carneiro, Brown and
  Suter}]{ransacTPS-eccv12}
\bibinfo{author}{Tran, Q.H.}, \bibinfo{author}{Chin, T.J.},
  \bibinfo{author}{Carneiro, G.}, \bibinfo{author}{Brown, M.S.},
  \bibinfo{author}{Suter, D.}, \bibinfo{year}{2012}.
\newblock \bibinfo{title}{In defence of ransac for outlier rejection in
  deformable registration}, in: \bibinfo{booktitle}{Proc. ECCV}.
\bibitem[{Varma and Ray(2007)}]{cit:varma2007}
\bibinfo{author}{Varma, M.}, \bibinfo{author}{Ray, D.}, \bibinfo{year}{2007}.
\newblock \bibinfo{title}{Learning the discriminative power-invariance
  trade-off}, in: \bibinfo{booktitle}{Proc. ICCV}.
\bibitem[{Vasconcelos et~al.(2017)Vasconcelos, Nascimento and
  Campos}]{vasconcelos2017prl}
\bibinfo{author}{Vasconcelos, L.O.}, \bibinfo{author}{Nascimento, E.R.},
  \bibinfo{author}{Campos, M.F.M.}, \bibinfo{year}{2017}.
\newblock \bibinfo{title}{{KVD}: Scale invariant keypoints by combining visual
  and depth data}.
\newblock \bibinfo{journal}{Pattern Recognition Letters} \bibinfo{volume}{86},
  \bibinfo{pages}{83 -- 89}.
\bibitem[{Vassileios~Balntas and Mikolajczyk(2016)}]{tfeat-balntas2016}
\bibinfo{author}{Vassileios~Balntas, Edgar~Riba, D.P.},
  \bibinfo{author}{Mikolajczyk, K.}, \bibinfo{year}{2016}.
\newblock \bibinfo{title}{Learning local feature descriptors with triplets and
  shallow convolutional neural networks}, in: \bibinfo{booktitle}{Proc. BMVC}.
\bibitem[{Wang et~al.(2019a)Wang, Ling, Lang, Feng and
  Hou}]{wang2019deformable}
\bibinfo{author}{Wang, T.}, \bibinfo{author}{Ling, H.}, \bibinfo{author}{Lang,
  C.}, \bibinfo{author}{Feng, S.}, \bibinfo{author}{Hou, X.},
  \bibinfo{year}{2019}a.
\newblock \bibinfo{title}{Deformable surface tracking by graph matching}, in:
  \bibinfo{booktitle}{Proc. ICCV}.
\bibitem[{Wang et~al.(2019b)Wang, Guo, Yan, Wang and Zhang}]{wang-cvpr19}
\bibinfo{author}{Wang, Y.}, \bibinfo{author}{Guo, J.}, \bibinfo{author}{Yan,
  D.M.}, \bibinfo{author}{Wang, K.}, \bibinfo{author}{Zhang, X.},
  \bibinfo{year}{2019}b.
\newblock \bibinfo{title}{A robust local spectral descriptor for matching
  non-rigid shapes with incompatible shape structures}, in:
  \bibinfo{booktitle}{Proc. CVPR}.
\bibitem[{Wilson and Snavely(2014)}]{cit:1dsfm}
\bibinfo{author}{Wilson, K.}, \bibinfo{author}{Snavely, N.},
  \bibinfo{year}{2014}.
\newblock \bibinfo{title}{Robust global translations with 1dsfm}, in:
  \bibinfo{booktitle}{Proc. ECCV}.
\bibitem[{Winder and Brown(2007)}]{cit:sift_learned_winder}
\bibinfo{author}{Winder, S.A.}, \bibinfo{author}{Brown, M.},
  \bibinfo{year}{2007}.
\newblock \bibinfo{title}{Learning local image descriptors}, in:
  \bibinfo{booktitle}{Proc. CVPR}.
\bibitem[{Wu et~al.(2020)Wu, Lin, Ding, Ni and Han}]{cit:wu2020aggregation}
\bibinfo{author}{Wu, G.}, \bibinfo{author}{Lin, Z.}, \bibinfo{author}{Ding,
  G.}, \bibinfo{author}{Ni, Q.}, \bibinfo{author}{Han, J.},
  \bibinfo{year}{2020}.
\newblock \bibinfo{title}{On aggregation of unsupervised deep binary descriptor
  with weak bits}.
\newblock \bibinfo{journal}{IEEE Trans. Image Process.} \bibinfo{volume}{29},
  \bibinfo{pages}{9266--9278}.
\bibitem[{Yi et~al.(2016)Yi, Trulls, Lepetit and Fua}]{cit:LIFT}
\bibinfo{author}{Yi, K.M.}, \bibinfo{author}{Trulls, E.},
  \bibinfo{author}{Lepetit, V.}, \bibinfo{author}{Fua, P.},
  \bibinfo{year}{2016}.
\newblock \bibinfo{title}{Lift: Learned invariant feature transform}, in:
  \bibinfo{booktitle}{Proc. ECCV}.
\bibitem[{Yu et~al.(2018)Yu, Liang, Xiao, Lu and Zheng}]{liang2018cviu}
\bibinfo{author}{Yu, Q.}, \bibinfo{author}{Liang, J.}, \bibinfo{author}{Xiao,
  J.}, \bibinfo{author}{Lu, H.}, \bibinfo{author}{Zheng, Z.},
  \bibinfo{year}{2018}.
\newblock \bibinfo{title}{A novel perspective invariant feature transform for
  rgb-d images}.
\newblock \bibinfo{journal}{Computer Vision and Image Understanding}
  \bibinfo{volume}{167}, \bibinfo{pages}{109 -- 120}.
\bibitem[{Zaharescu et~al.(2009)Zaharescu, Boyer, Varanasi and
  Horaud}]{zaharescu2009cvpr}
\bibinfo{author}{Zaharescu, A.}, \bibinfo{author}{Boyer, E.},
  \bibinfo{author}{Varanasi, K.}, \bibinfo{author}{Horaud, R.P.},
  \bibinfo{year}{2009}.
\newblock \bibinfo{title}{{Surface Feature Detection and Description with
  Applications to Mesh Matching}}, in: \bibinfo{booktitle}{Proc. CVPR}.
\bibitem[{Zeng et~al.(2017)Zeng, Song, Nie{\ss}ner, Fisher, Xiao and
  Funkhouser}]{3dmatch-cvpr17}
\bibinfo{author}{Zeng, A.}, \bibinfo{author}{Song, S.},
  \bibinfo{author}{Nie{\ss}ner, M.}, \bibinfo{author}{Fisher, M.},
  \bibinfo{author}{Xiao, J.}, \bibinfo{author}{Funkhouser, T.},
  \bibinfo{year}{2017}.
\newblock \bibinfo{title}{{3DMatch}: Learning local geometric descriptors from
  rgb-d reconstructions}, in: \bibinfo{booktitle}{Proc. CVPR}.
\bibitem[{Zhang et~al.(2018)Zhang, Isola, Efros, Shechtman and
  Wang}]{zhang2018perceptual}
\bibinfo{author}{Zhang, R.}, \bibinfo{author}{Isola, P.},
  \bibinfo{author}{Efros, A.A.}, \bibinfo{author}{Shechtman, E.},
  \bibinfo{author}{Wang, O.}, \bibinfo{year}{2018}.
\newblock \bibinfo{title}{The unreasonable effectiveness of deep features as a
  perceptual metric}, in: \bibinfo{booktitle}{Proc. CVPR}.
\bibitem[{Zhao et~al.(2015)Zhao, Feng, Wan and Zhang}]{sphorb}
\bibinfo{author}{Zhao, Q.}, \bibinfo{author}{Feng, W.}, \bibinfo{author}{Wan,
  L.}, \bibinfo{author}{Zhang, J.}, \bibinfo{year}{2015}.
\newblock \bibinfo{title}{Sphorb: A fast and robust binary feature on the
  sphere}.
\newblock \bibinfo{journal}{Int. J. Comput. Vis.} \bibinfo{volume}{113}.

\end{thebibliography}
